\documentclass[10pt,twocolumn,letterpaper]{article}

\usepackage[pagenumbers]{cvpr} % To force page numbers, e.g. for an arXiv version
\usepackage{float}

\usepackage{amsthm}

\theoremstyle{definition}

\theoremstyle{remark}

\definecolor{cvprblue}{rgb}{0.21,0.49,0.74}
\usepackage[pagebackref,breaklinks,colorlinks,allcolors=cvprblue]{hyperref}
\usepackage{subcaption}
\usepackage{svg}
\usepackage{caption}
\usepackage[accsupp]{axessibility}

\def\paperID{11799} % *** Enter the Paper ID here
\def\confName{CVPR}
\def\confYear{2026}

\title{A combination of 
noise and bilateral filters achieve supralinear and scalable adversarial robustness in CNNs}
\author{Nicolas Stalder\\
{\tt\small nix.stalder@hotmail.com}
\and
Benjamin F. Grewe\\
{\tt\small bgrewe@ethz.ch}
\and
Matteo Saponati$^*$\\
{\tt\small masapo@ini.ethz.ch}
\and
Pau Vilimelis Aceituno\thanks{M.S. and P.V.A. contributed equally as senior authors} \\
{\tt\small pau@ini.ethz.ch}\\
Institute of Neuroinformatics \\ETH Zürich, University of Zürich \\
Winterthurerstrasse 190, 8057 Zürich, Switzerland\\
}

\begin{document}
\maketitle
\begin{abstract}

    The vulnerability of deep neural networks to adversarial examples poses a significant challenge for real-world deployment. Existing techniques to enhance deep network robustness rely on adversarial training, an approach that is powerful but computationally intensive and typically tailored to specific attack types. 
    To address these limitations, existing works have explored techniques such as adding gaussian noise or filtering images, both of which can boost the network robustness to various adversarial attacks, albeit modestly. Here, we theoretically demonstrate that these two approaches enhance robustness against adversarial attacks through complementary mechanisms, resulting in supralinear robustness when combined.
    Building on this insight, we experimentally show that a simple preprocessor combining Gaussian noise and bilateral filtering yields supralinear improvements in adversarial robustness with minimal computational cost.
    Next, we combine our preprocessor with adversarial training and test on RobustBench to assess its supralinear improvement over state-of-the-art defenses. First, this combination ranks second on AutoAttack and third overall, while using only $\sim$35\% of the training FLOPs, using a model with 50\% less parametets, trained with $\sim$33\% of the epochs and $\sim$15\% the data compared to state-of-the-art defenses.
    Second, our method scales efficiently, matching the accuracy of competing models with roughly 2–8× less total compute across $\sim$3 orders of magnitude. 
    Overall, our approach provides a principled and easily integrable framework for enhancing adversarial robustness, offering negligible computational overhead and a simple yet theoretically grounded design.
\end{abstract}

\section{Introduction}
A decade ago, Convolutional Neural Networks (CNNs) achieved superhuman performance in object recognition tasks \cite{ciresan_high-performance_2011, krizhevsky_imagenet_2012, he_deep_2015}.
However, targeted pixel-level perturbations of the original image can cause CNNs to make incorrect predictions \cite{bruna_intriguing_2014, goodfellow_explaining_2015}, while being often imperceptible to humans \cite{chen2023diffusionmodelsimperceptibletransferable}.
These adversarial attacks (AAs) present a significant challenge when CNNs are used to solve real-world problems, for example, self-driving systems, \cite{pavlitska2023adversarial}, facial recognition \cite{vakhshiteh2021adversarial, khan_facial_2019}, and many others. 

Adversarial training remains one of the most effective defenses against such attacks \cite{goodfellow_explaining_2015,wang_better_2023}, but it is computationally expensive, needing around 9x more floating point operations (FLOPs) \cite{bartoldson_adversarial_2024} and therefore scaling poorly with model size.
Additionally, its robustness is tied to attacks seen during training \cite{tramèr2019adversarialtrainingrobustnessmultiple}, which can lead to difficulties when deploy in real-world scenarios.
In contrast, simple filtering techniques or image corruptions using additive noise can improve robustness to certain adversarial attacks with minimal computational cost \cite{ziyadinov_low-pass_2023}.
However, no single filtering or image corruption method has proven broadly effective across various attack types (see Section \ref{sec:related-work} for details).
Importantly, these approaches have largely been developed in isolation, and a systematic evaluation of their combined effect on adversarial robustness is still lacking.

Motivated by these observations, we asked whether a simple, theoretically grounded combination of filtering and additive noise could enhance adversarial robustness while easily integrating with existing defense methods.
We address this question both mathematically and with extensive experiments with both standard CNNs and state-of-the-art models, making the following contributions:
\begin{enumerate}
\item We show mathematically that Gaussian noise and filtering enhance adversarial robustness through distinct mechanisms, and their combination protects against a broad class of attacks (Section~\ref{sec:theory}).
\item We verify this result empirically, showing that Gaussian noise and filters can yield a supralinear gain in adversarial robustness (Section \ref{sec:results-adversarial-attacks}).
\item When combined with adversarial training, we achieve \textbf{+0.6\%} higher robust accuracy compared to the previous state-of-the-art, while using only $\sim$\textbf{35\%} of the training FLOPs (Section \ref{sec:results-SOTA}).
\item Our preprocessor scales efficiently across model sizes, reducing the training compute needed to reach competitive robustness by up to \textbf{15×} (Section \ref{sec:results-SOTA})..
\end{enumerate}
\section{Related work}
\label{sec:related-work}
We review existing defense mechanisms against adversarial attacks, which we group these methods into three categories that are particularly relevant to this work. 

\textbf{Introducing artificial samples in the training dataset}. 
Introducing artificial samples into the training data is a commonly used strategy in machine learning, and also includes the most prominent adversarial defense method.
The most prominent approach integrates adversarial examples into the training process \cite{goodfellow_explaining_2015, madry_towards_2017, wong_fast_2020}, generally referred to as adversarial training.
These defenses, when combined with synthetic data generated by diffusion models \cite{wang_better_2023, bartoldson_adversarial_2024}, currently represent the SotA on visual datasets \cite{croce2020robustbench}.
However, their applicability in real-world scenarios is limited due to the high computational cost of adversarial training, which requires increasingly large and better artificial datasets to maintain effectiveness \cite{bartoldson_adversarial_2024}.
Moreover, they may fail to defend against perturbations not encountered during training and in some cases, even increase susceptibility to such attacks \cite{tramèr2019adversarialtrainingrobustnessmultiple}, necessitating costly retraining whenever new attack types emerge.
Noise addition during training also belongs to this category.
Simpler techniques such as Gaussian, Poisson, or uniform noise addition during training can improve robustness to adversarial attacks \cite{shi_defending_2022, ford2019adversarialexamplesnaturalconsequence}, while also enhancing a network’s resilience to noise at inference time \cite{shi_defending_2022, li_certified_2019, cohen_certified_2019}.

\textbf{Additive noise and random transformation methods}. 
The second category encompasses methods that introduce randomness at inference time to inhibit adversarial attacks by making gradients unreliable.
This randomness can be implemented via various techniques: noise addition to input data \cite{shi_defending_2022}, noise addition within the network \cite{li_certified_2019, cohen_certified_2019}, random input transformations such as resizing \cite{xie2018mitigatingadversarialeffectsrandomization}, or applying a randomly selected transformation from a set \cite{Raff_2019_CVPR}.
To mitigate the accuracy degradation introduced by randomness, these methods often incorporate additional mechanisms.
First, they train models with noise to improve the inference accuracy under noise addition \cite{shi_defending_2022, li_certified_2019, cohen_certified_2019}.
Second, they employ ensembles of stochastic models, which classify samples via majority vote \cite{cohen_certified_2019, li_certified_2019, LecyuerDifferntial}.
However, ensemble methods drastically increase computational cost during inference, limiting their practicality in real-time or resource-constrained settings.

\textbf{Filtering methods}. 
In the third category, adversarial perturbations are filtered out through various means.
These include simple methods like Gaussian filtering \cite{ziyadinov_low-pass_2023} and bilateral filtering \cite{tomasi_bilateral_1998, ratzlaff_unifying_2018}, as well as more complex image processing strategies such as JPEG compression \cite{guo_countering_2017, dziugaite_study_2016, das_keeping_2017}.
Another line of work employs denoising neural networks (NNs) \cite{liao_defense_2018, gu_towards_2015}, which can enhance robustness but come with significant computational overhead during both training and inference.
A further key limitation of NN-based preprocessors is their own vulnerability to adversarial attacks, which can undermine their defensive effectiveness \cite{gu_towards_2015}.
More recent work explores combining noise addition with denoising NNs \cite{carlini2023certifiedadversarialrobustnessfree}.
This approach leverages the synergy between added noise and subsequent denoising to improve robust accuracy, as the denoiser simultaneously attenuates adversarial perturbations and filters out the introduced noise.

\textbf{Combining filtering and additive noise.}
Our preprocessor combines  bilateral filtering (a filtering methods) and Gaussian noise addition (a type of additive noise perturbation) during both training and inference.
Thus, our approach is effectively a combination of the three defense mechanisms described above.
To the best of our knowledge, such combination has not yet been explored, nor has the joint use of any simple noise and filtering methods applied during both training and inference.
\section{Our Preprocessor}
\label{sec:preprocessor}

We preprocess the input data with a combination of independent zero-mean Gaussian noise to each pixel channel and multiple iterations of bilateral filtering.
We chose pixel-wise Gaussian noise due to its simplicity and demonstrated effectiveness in improving adversarial robustness \cite{shi_defending_2022, li_certified_2019, cohen_certified_2019}.
For the filtering step, we found that repeatedly applying bilateral filters gave the best results, as they effectively reduce noise while preserving sharp edges and image structure \cite{tomasi_bilateral_1998}.
See Appendix \ref{sec:filtering-and-noise-methods} for a detailed description of the two components.
\section{Theory}
\label{sec:theory}

Adversarial attacks in the context of image classification are small pixel-wise perturbations of images that change the classification provided by a deep neuronal network. Such attacks are different from random perturbations (i.e. noise) or coarse modifications (i.e. image counterfeiting), and are often input-specific found by sophisticated optimization procedures. Furthermore, those attacks are generally imperceptible to humans, a feature that makes them particularly problematic for real-world deployment. In this section we present our theory referring to both attacks and potential defenses, with a detailed analysis presented in Appendix \ref{sec:math}. 

\begin{figure}[ht!]
    \centering
\includegraphics[width = 1.\linewidth]{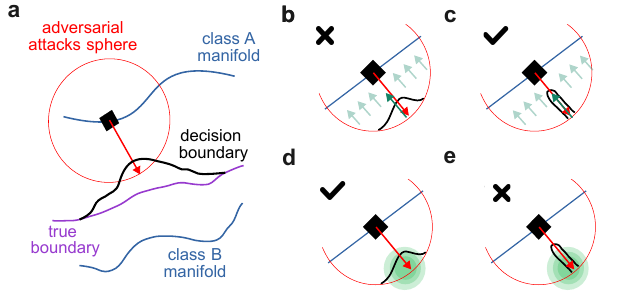}
\caption{
Illustration of the mathematical analysis.
(a) Concepts used in our analysis: An image (black rhombus) lies on a class manifold (blue line). The neural network separates these classes with a decision boundary (black line). An adversarial attack (red arrow) is a small perturbation of the image within a specified radius (red circle) that pushes it across this boundary.
(\textbf{b}-\textbf{c}) Adversarial attacks in the presence of a filter (green arrows). The filter attenuates perturbations from the image manifold, including those introduced by adversarial attacks. In (\textbf{b}), the adversarial perturbation lies far from the image manifold, so the filter effectively cancels it. In (\textbf{c}), the attack is close to the image manifold and cannot be fully suppressed by the filter. (\textbf{d}-\textbf{e}) Adversarial attacks against noise (green area). Noise introduces random deviations unknown to the attacker, which can disrupt adversarial perturbations. The attack that fails against the filter can succeed with high probability in the presence of noise, as the noise covers a broad region around the attack (\textbf{d}). Conversely, the attack that succeeds against the filter is likely to fail when noise is added, since it is not spatially clustered (\textbf{e}).
}
\label{fig:mathIllustration}
\end{figure}

We start by providing some basic definitions which are illustrated in Fig~\ref{fig:mathIllustration}. We consider a set of input images $x\in \mathcal{X}$, where $\mathcal{X}$ is the space of possible images with dimension $D$. Unaltered images belong to a single class, which is defined by a set of points in image space. A neural network $f(\cdot)$ receives an input image and classifies it as belonging to one of the classes. The neural network can be characterized by a set of decision boundaries between pairs of classes in image space which are smooth $D-1$ dimensional manifolds that separate two classes, with a maximum sectional curvature $c$. 

An adversarial attack for an image $x$ that is (correctly) classified as belonging to a given class is a vector $a_x$ such that $x+a_x$ crosses the decision boundary. However, not all vectors are valid: an adversarial attack must have a small norm $\|a_x\|\leq r$, which defines a sphere around the image that we will denote as an adversarial sphere. Furthermore, it must be not generate a misclassifications by a human (otherwise it would be image counterfeiting). Thus, we assume the existence of a unique human classification function $h_{c,c'}$ which defines a "true" decision boundary. This boundary allows us to define an adversarial volume reflecting the amount of adversarial attacks,
\begin{equation}
    V_a(x) = \int_{z\in \mathbf{B}(x,r)} \mathbf{1}\left[f(z)\neq h(z)=f(x)\right]dz
\end{equation}
where $\mathbf{B}(x,r)$ is a ball of radius $r$ around $x$. $V_a(x)$ is a centerpiece of our analysis, as it allows us to compare adversarial defenses: if we fix a given $V_a(x)$, we can then ask what is the shape of adversarial attacks that would bypass a given defense.  Furthermore, it is also relevant for adversarial training, as existing theoretical works argue that adversarial training works by reducing "dimples" in the decision boundary, which is equivalent to reducing $V_a(x)$ \cite{shamir2021dimpled}. Thus, our analysis here will be a natural complement to adversarial training.

\textbf{Filters: } To analyze filters, we consider an idealized de-noising filter which we model as a function $\varphi(x):\mathcal{X}\rightarrow\mathcal{X}$ that leaves the unaltered images unaffected, attenuates any deviation from the image, and is otherwise smooth and has full range. 

Intuitively, an idealized filter pushes any perturbation of the image (such as an adversarial attack) back towards the image manifold. In the worst case, the decision boundary would be very close to the image manifold, and thus the filter would not be able to push the perturbed image back to the correct side. Note however, that this setting only requires one point to be close to the boundary, and thus we only need to allocate the adversarial volume $V_a(x)$ accordingly. See Fig.~\ref{fig:mathIllustration}.b-c for a best- and worst-case boundary with a fixed $V_a(x)$.

More precisely, the worst possible shape of the decision boundary is a thin cylinder that enters the adversarial sphere and gets as close as possible to the image. Thus, for a fixed  adversarial volume $V_a(x)$, there could be a successful adversarial attack if
\begin{equation}
    V_a(x) >r\left(1-\lambda_{\varphi(x)}^{\min}\right)S_{D-1}(c^{-1/2}),
\end{equation}
where $\lambda_{\varphi(x)}^{\min}$ is the minimum eigenvalue of the Jacobian matrix $J_{\varphi}$ of $\varphi$, and $S_{D-1}(c^{-1/2})$ is the volume of a sphere of dimension $D-1$ with curvature $c$ (inverse of the squared radius). And even if such an attack exists, the adversarial volume will be reduced by a factor of $|J_\varphi|\frac{r-\alpha}{r}$, where $\alpha$ is the ratio between $V_a(x)$ and $S_{D-1}(c^{-1/2})$. The proofs are presented in Appendix \ref{app:math_attackAgainstFilter}.

\textbf{Noise: }To study noise, we consider an image $x$ perturbed by an optimal adversarial attack $a_x$ and noise $\varepsilon$ drawn from a Gaussian distribution with variance $\sigma^2$. Intuitively, the noise moves a perturbed image in a random direction, and thus it would cross the boundary if that boundary covers a low fraction of the volume around the perturbed image. Thus, the worst-case scenario is a decision boundary that is spherical around the attack, and the best case is a thin filament. See Fig.~\ref{fig:mathIllustration}.d-e for a best- and worst-case boundary with a fixed $V_a(x)$.

Thus, the worst possible arrangement of the adversarial volume is given by a sphere placed at the edge of the adversarial sphere, which results in the following upper bound on the probability of  an attack succeeding 
\begin{equation}\label{eq:boundOnNOise}
    \text{Pr}\left[x+a_x+\varepsilon\in A \right] \leq 1-2\text{erf}\left(-\frac{\rho_D(V_a(x))}{\sigma}\right),
\end{equation}
where erf is the cumulative gaussian function, and $\rho_D(V_a(x))$ is the radius of a sphere of volume $V_a(x)$ in dimension $D$. The proof is presented in Appendix \ref{app:math_attackAgainstNoise}. We extended the probabilistic bound to a combination of filters and noise in the Appendix \ref{app:math_combination}.

A crucial point of our theory  is that both defenses target fundamentally different shapes of attacks (see Fig.~\ref{fig:mathIllustration}). Gaussian noise works when adversarial attacks that are distributed in thin filaments around the image, while it fails if the attacks are clustered into large spheres in image space. In contrast, filters work well when attacks are clustered close to the edge of the maximum allowed norm of the adversarial attack, but fail when they are close to the image. Thus, our analysis suggest that combining those methods will cover a wider range of potential attacks than using either method alone. 

Before testing out theory, it is worth making two remarks. First, our analysis focused on treating noise and filters separately, ignoring their interactions. Analyzing the combined effects of filters suggests that the effects are non-commutative, but it is not clear which order would result in higher accuracy; yet, we expect that adding noise and then filtering should improve the system's clean accuracy (see Appendix \ref{app:math_combination}), making this our default recommendation. Second, we would expect that adversarial training pushes the learned boundary towards the true boundary, thus reducing $V_a(x)$ \cite{shamir2021dimpled}. This it constitutes yet a different approach that could further compound gains, albeit with significant training overhead.

\section{Experiments}
\begin{table*}[!ht]
\caption{
    Ablation study table, showing test accuracy (in \%) on the CIFAR10 dataset and under different adversarial attacks.
    We test standard CNN models \cite{tan_efficientnet_2020} trained without (first row) and with different defense methods.
    We also report sum of the accuracy gain of the individual methods per attack as linear gain and the actual gain when combining does two( Noise + Bil.) as actual gain.
    For \(L_{\infty}\) attacks, we set \(\epsilon = 0.03 \approx 8/255\), and for \(L_{2}\) attacks, we set \(\epsilon = 0.51\), which are standard values in the literature \cite{das_keeping_2017,dziugaite_study_2016}.
    For the C\&W attack, we set \(c = 1.8\)  for the standard CNN as the strongest tested attacks.
    Hyperparameters for all methods and the adversarial attacks are listed in the appendix Tables \ref{tabel:preproc-hyperparams} and \ref{tabel:adv-hyperparams}, respectively.
}

\label{table:supralinear-table}
\vspace{11pt}
\centering
\resizebox{0.8\linewidth}{!}{%
\renewcommand{\arraystretch}{1.2}
{
\begin{tabular}{lllllll}
\toprule\toprule
Method   & Clean & FGSM     &  \(L_{\infty}\) & EoT.  & \(L_{2}\)& C\&W \\
% \bottomrule
%
% (result 1)
%
%
% & & & & & & &\\[1pt]
\bottomrule\bottomrule
Standard CNN \cite{tan_efficientnet_2020} &74.5\% \(\pm\)2.4  & 3.5\% \(\pm\)2.1     & 0.2\% \(\pm\)0.4     & 0.2\% \(\pm\)0.5        & 1.3\% \(\pm\)0.6    & 0.6\% \(\pm\)0.6       \\          
\midrule
+ Bil.  &69.0\%  \(\pm\)2.2\  &10.0\% \(\pm\)1.2     & 1.0\% \(\pm\)0.4    & 1.2\% \(\pm\)0.5        &11.9\% \(\pm\)1.4    & 0.5\% \(\pm\)0.3     \\
+ Noise       &68.5\% \(\pm\)1.7\  &22.8\% \(\pm\)1.3     &25.5\% \(\pm\)1.2    &12.0\% \(\pm\)0.9         &49.6\% \(\pm\)1.4    &43.0\% \(\pm\)1.0   \\
+ Noise + Bil. &67.9\% \(\pm\)1.4\  &33.9\% \(\pm\)1.7     &36.5\% \(\pm\)1.5    &18.9\% \(\pm\)1.4        &58.5\% \(\pm\)1.7    &47.2\% \(\pm\)1.0     \\
+ Bil. + Noise &67.5\% \(\pm\)0.3\  &30.8\% \(\pm\)0.5     &28.1\% \(\pm\)0.5    &15.1\% \(\pm\)0.5        &54.1\% \(\pm\)0.4    &41.8\% \(\pm\)0.6     \\
\midrule
linear gain & -11.5\% &25.8\%  &26.1\%  &12.8\%    &\textbf{58.9}\%    &42.3\%   \\
actual gain & \textbf{-6.6}\% &\textbf{30.4}\%  &\textbf{36.2}\%  &\textbf{18.7}\%    &57.2\%    &\textbf{46.6}\%   \\
\bottomrule
\end{tabular}
}}

\end{table*}
We mostly used the CIFAR-10 dataset \cite{cifar_10} in our experiments.
We further use the Imagenet10 dataset \cite{liu2020imagenet10} in a short ablation study (see Appendix \ref{sec:imagenet10}) to show that our preprocessor also generalizes to other datasets.

Here, we summarize the key experimental details.
If not specified otherwise we apply noise first in our preprocessor.
In Section \ref{sec:results-adversarial-attacks}, we used the EfficientNet-B0 \cite{tan_efficientnet_2020} as a representative standard CNN, trained for 80 epochs.
In Section \ref{sec:results-SOTA}, we used the Wide Residual Network (WRN) model of different sizes \cite{zagoruyko_wide_2017, hendrycks_gaussian_2023} with Swish activation functions \cite{hendrycks_gaussian_2023}.
For the WRN training, we follow the adversarial training protocol introduced in TRADES \cite{zhang_theoretically_2019}, using the setup from \cite{wang_better_2023}. 
In addition, we augment training data with synthetic samples generated by an elucidating diffusion model (EDM) \cite{Karras2022edm}. 
The amount of generated data varies depending on the number of training epochs.
Due to the high computational cost of adversarial training, we were unable to report standard deviations for experiments on the WRN. 
See Appendix~\ref{sec:experimental-details} for a comprehensive description.
\subsection{Ablation study}
\label{sec:results-adversarial-attacks}
In this first section, we perform an ablation study to demonstrate that both the bilateral filter \cite{tomasi_bilateral_1998} and the Gaussian noise component are necessary to achieve the highest robust accuracy with our preprocessing strategy.
To do so, we train standard CNN models \cite{tan_efficientnet_2020} with and without different pre-processing methods on the CIFAR-10 dataset.
Each model variant is evaluated against multiple adversarial attacks and Gaussian noise perturbations (see Appendix~\ref{sec:experimental-details} for a description of the different adversarial attacks).

Our ablation study (see Table \ref{table:supralinear-table}) shows that adding Gaussian noise during both training and inference increases robustness against all types of adversarial perturbations, whereas adding a bilateral filter during both phases improves robustness only against certain perturbations.
When these two methods are combined, they achieve higher accuracy than either component alone across all adversarial perturbations.
Comparing the accuracy gains of the preprocessor models relative to the standard CNN, we observe that for four out of six perturbations, the combined model’s gain exceeds the sum of the individual gains — a supralinear effect.
Under the \(L_{2}\) attack, where the combined gain is smaller than the sum of its parts, both individual preprocessors already show high robustness relative to their performance under other adversarial perturbations.
Regarding clean accuracy, all preprocessing variants show a reduction, with the combined preprocessor having the lowest clean accuracy.
Interestingly, the drop in clean accuracy for the combined model is less than the sum of the individual drops.

\begin{figure*}[ht!]
    \centering
     \begin{subfigure}[b]{0.9\textwidth}
        \centering
        \includegraphics[width=0.95\linewidth]{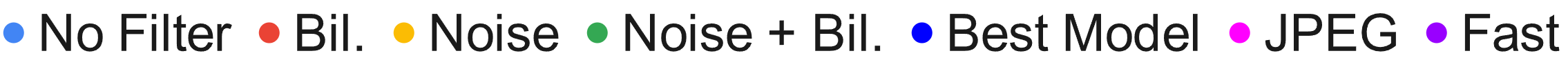}
    \end{subfigure}

    \vspace{0.1cm} % Add some vertical space between legend and charts
    
    % First row
    \begin{subfigure}[b]{0.45\textwidth}
        \centering
        \includegraphics[width=\linewidth]{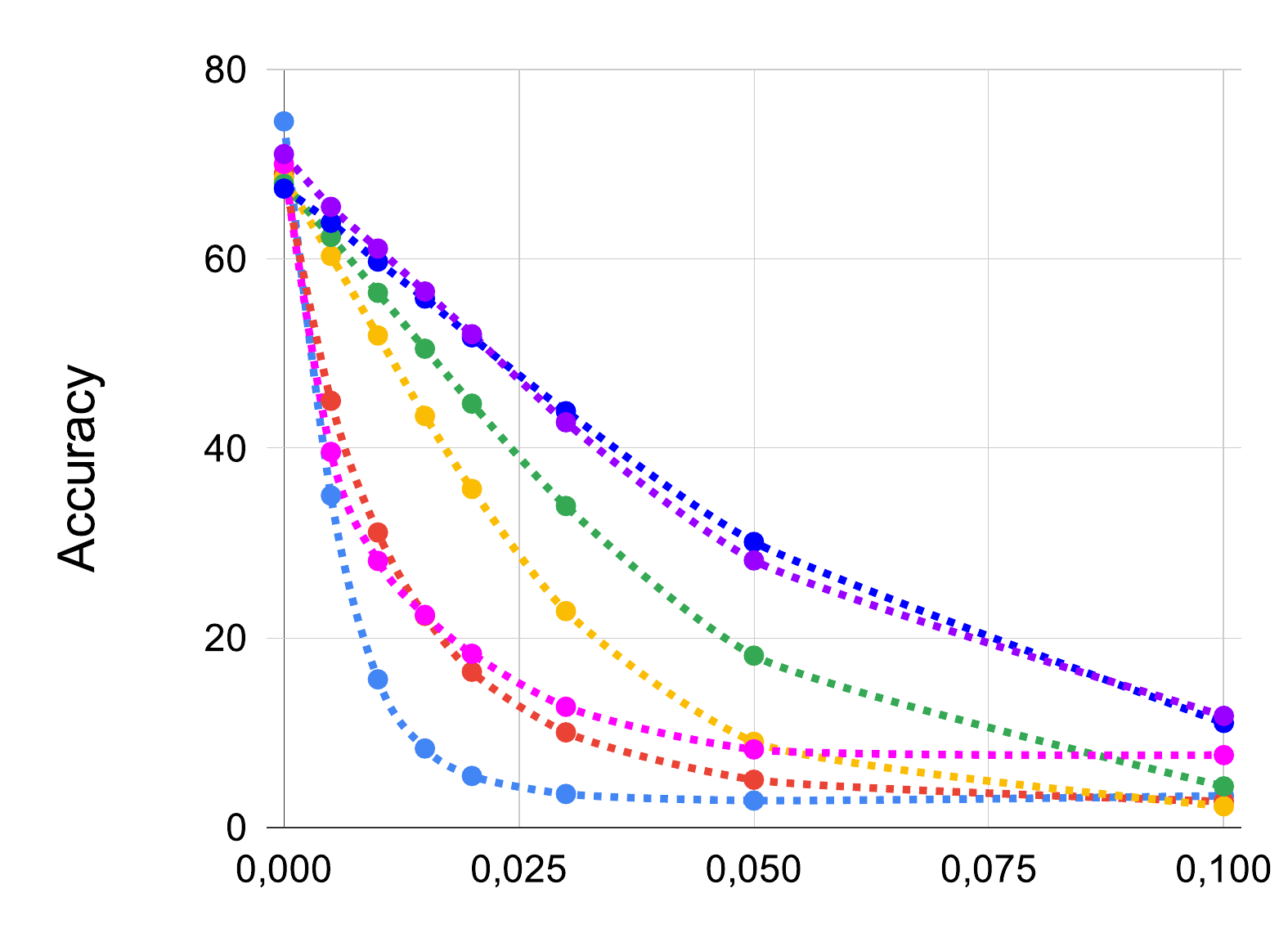}
        \caption{FGSM}
        \label{fig:basic_fgsm}
    \end{subfigure}
    \hfill
    \begin{subfigure}[b]{0.45\textwidth}
        \centering
        \includegraphics[width=\linewidth]{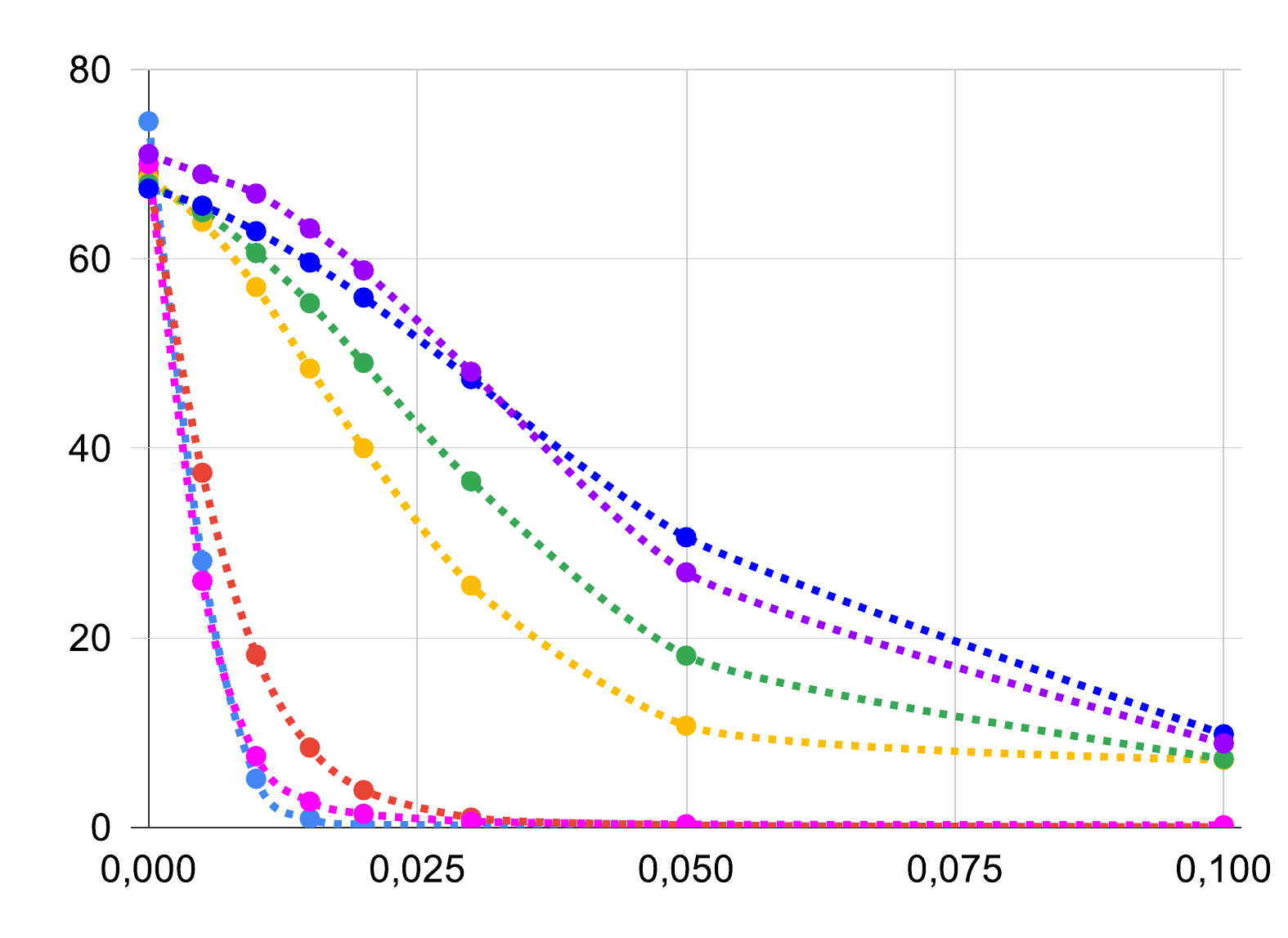}
        \caption{APGD-INF}
        \label{fig:basic_apgdinf}
    \end{subfigure}

    % Second row
    \begin{subfigure}[b]{0.45\textwidth}
        \centering
        \includegraphics[width=\linewidth]{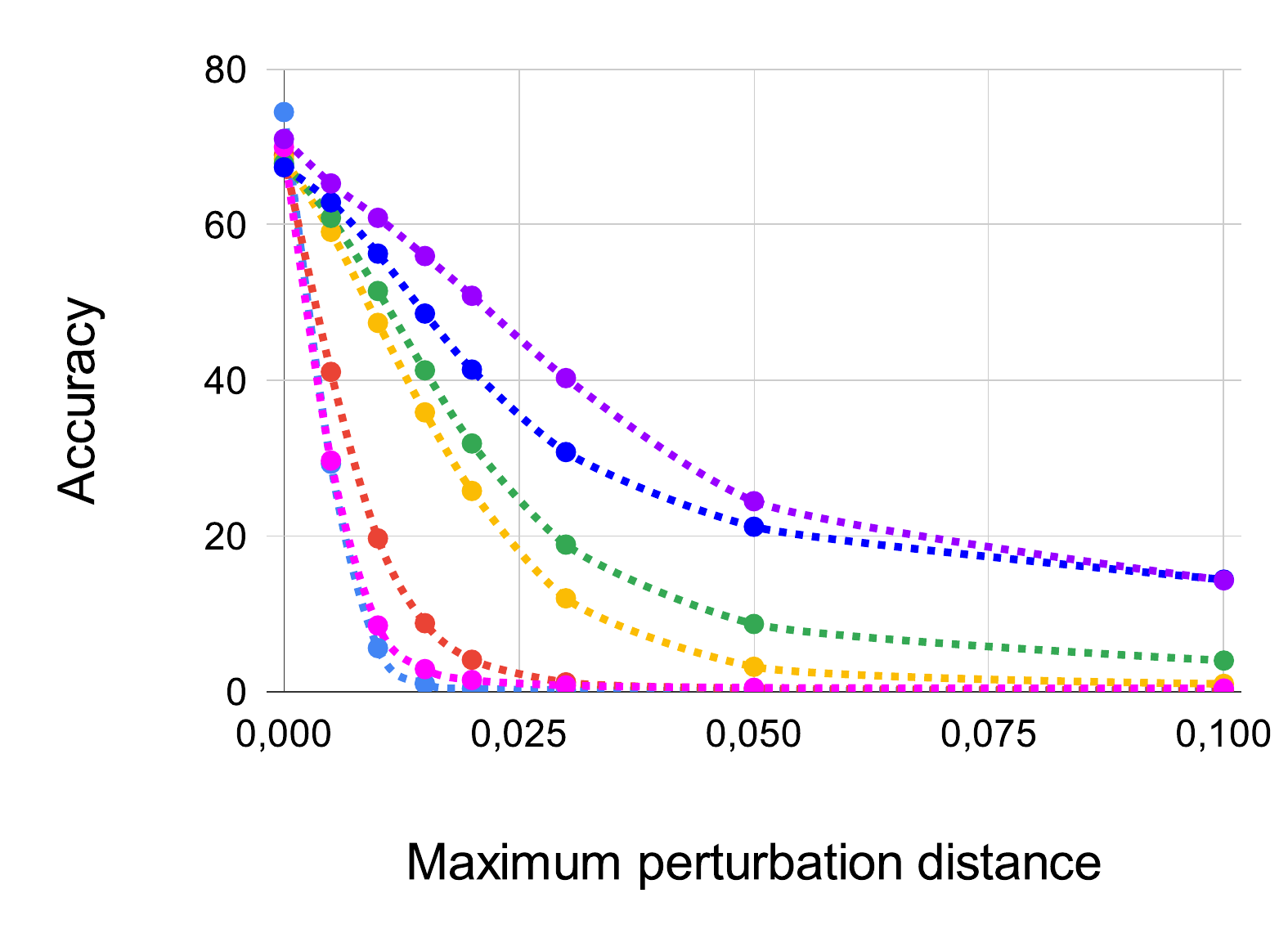}
        \caption{EoT}
        \label{fig:basic_eot}
    \end{subfigure}
    \hfill
    \begin{subfigure}[b]{0.45\textwidth}
        \centering
        \includegraphics[width=\linewidth]{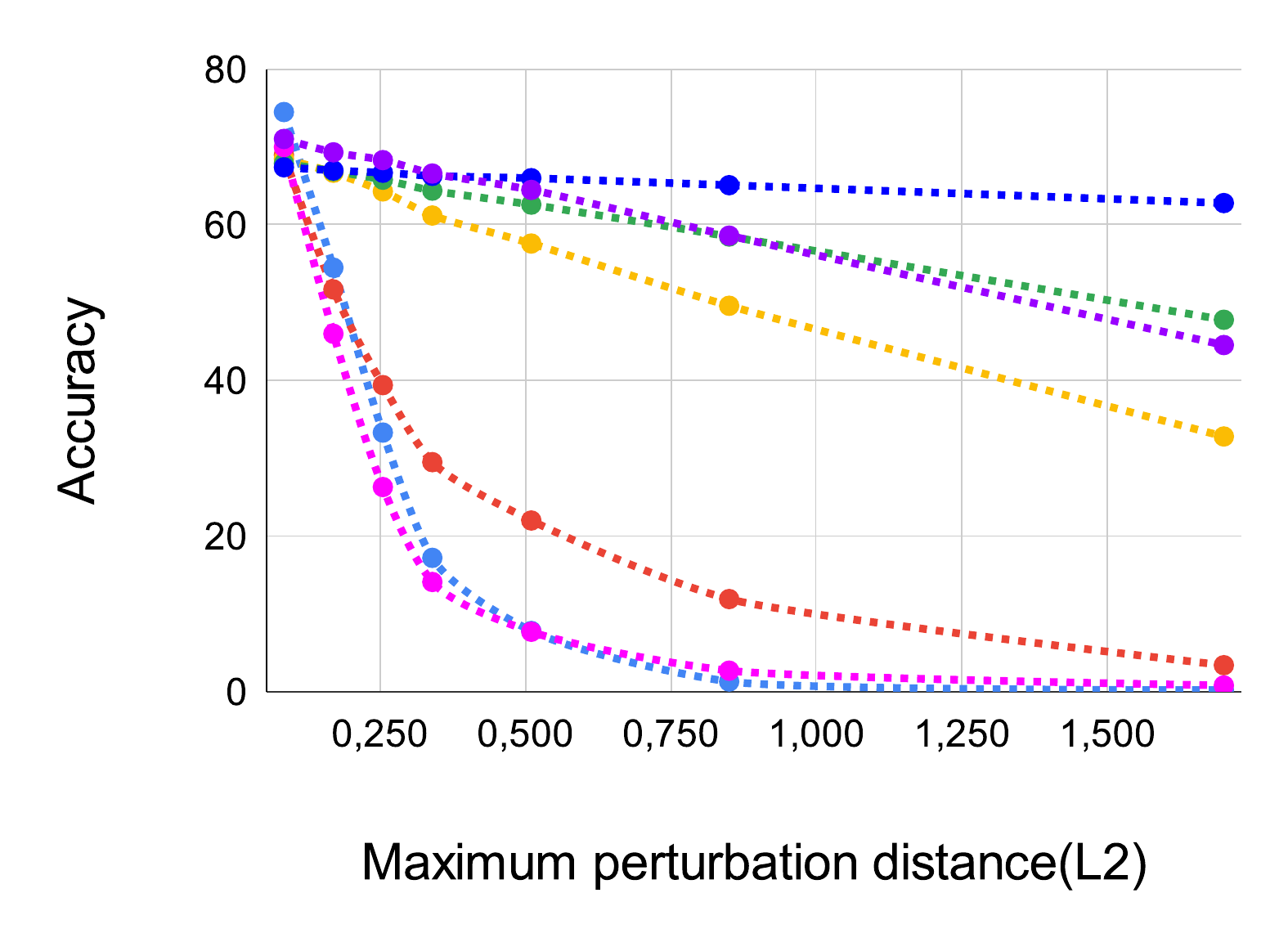}
        \caption{APGD2}
        \label{fig:basic_apgd2}
    \end{subfigure}

    \caption{Accuracy of different methods against adversarial attacks on the Standard CNN~\cite{tan_efficientnet_2020}, showing results for all tested $\epsilon$ values. 
    The experimental details are provided in Appendix~\ref{sec:experimental-details}.}
    \label{fig:basic}
\end{figure*}

\begin{figure*}[ht]
    \centering
    \includegraphics[width=\linewidth]{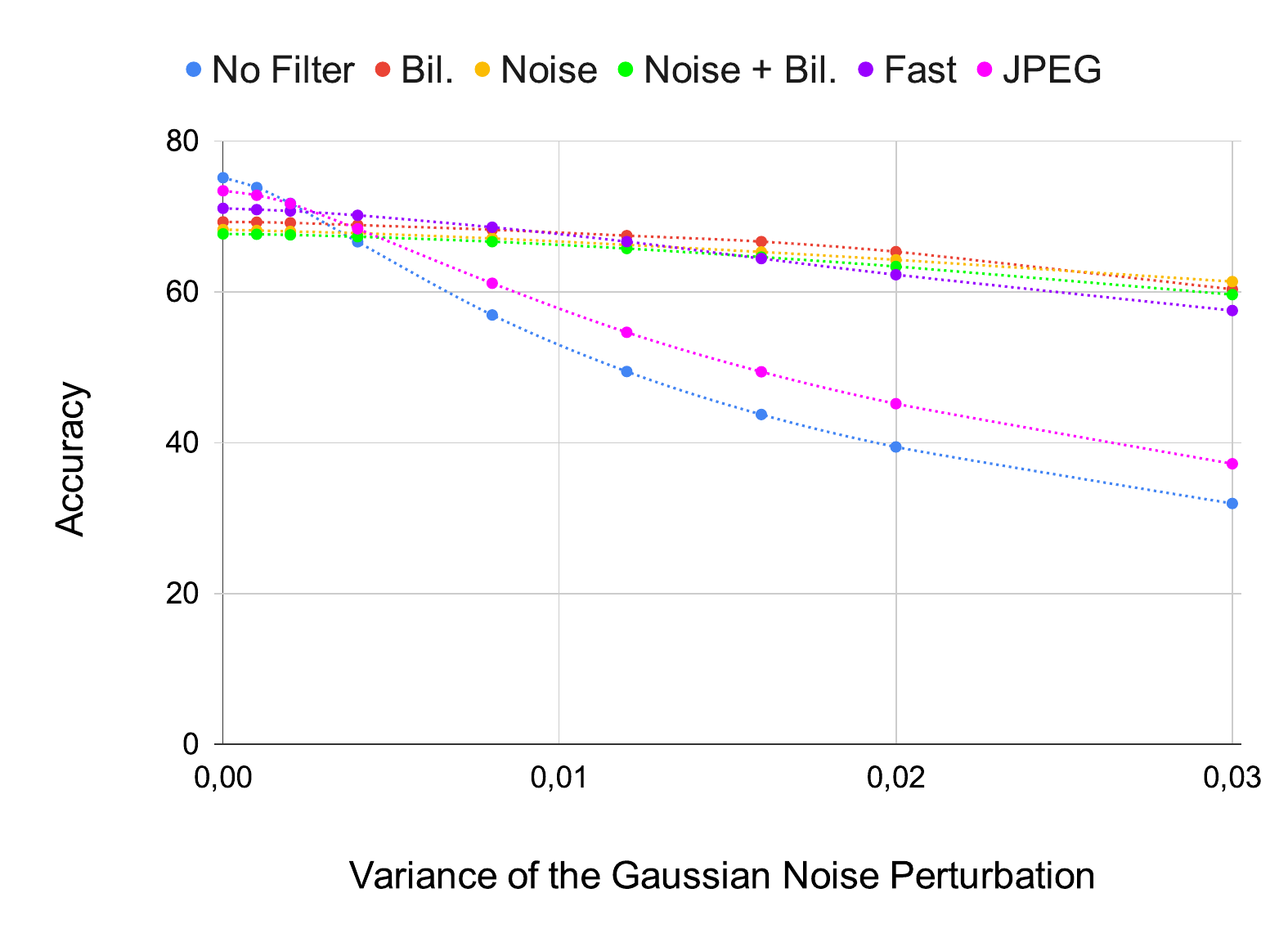}
    \caption{Preprocessors with the Standard CNN \cite{tan_efficientnet_2020} vs. Gaussian noise perturbations.The experimental details are listed in Appendix \ref{sec:experimental-details}.}
    \label{fig:basic_noise}
\end{figure*}

We further conducted experiments over a wider range of adversarial perturbation magnitudes and Gaussian noise intensities, and also evaluated several other defense methods: JPEG compression \cite{das_keeping_2017,dziugaite_study_2016,guo_countering_2017}, Fast Adversarial Training \cite{wong_fast_2020} and our best combined preprocessor (see Tables \ref{fig:basic} and \ref{fig:basic_noise}).
These results show that while both Fast Adversarial Training and our best preprocessor achieve strong robustness against most attacks and Gaussian noise, each exhibits specific weaknesses (see Table \ref{table:full-tabel}).
Fast Adversarial Training performs poorly against the C\&W attack \cite{carlini_towards_2017}, with a robust accuracy of 20.9\%, whereas our best preprocessor is weakest against the EoT attack \cite{zimmermann_comment_2019}, with a robust accuracy of 30.8\%.
JPEG compression is somewhat less effective than bilateral filtering alone for adversarial perturbations and is significantly less robust against Gaussian noise.
Finally, to determine suitable preprocessor parameters, we conducted an extensive sensitivity analysis using FGSM (see Figure \ref{fig:sensitivity}).
\begin{table*}[ht!]
\caption{
Test accuracy (in \%) on the clean CIFAR-10 dataset and under AutoAttack and its EoT variant.
\emph{Prepro. 10/50} and \emph{Prepro. 20/80} refer to the number of bilateral filters used during training and inference, respectively (e.g., 10 filters during training and 50 during inference). 
The preprocessor also includes Gaussian noise with zero mean and a variance of 0.032.
Our best model is a WRN-82-12 with a 20/100 preprocessor, while the current state-of-the-art (SotA) model without postprocessing is a WRN-94-16.
All WRN-28-4, 28-10 and our best model in this table are trained and evaluated by us, while the WRN-82-12 without preprocessor and the 94-16 are taken from the literature \cite{bartoldson_adversarial_2024} for a comparison.
The number of epochs of adversarial training varies based on the amount of generated data and the model used. 
Best results under the same training condition as well as the best results comparing the SotA, WRN-82-12 and our best model are highlighted in bold.
Hyperparameters for the preprocessor and the adversarial training are listed in the appendix Tables \ref{table:WRN-preproc} and \ref{table:WRN-train}, respectively.
}
\label{table:auto-attack}
\vspace{11pt}
\centering
\resizebox{0.7\linewidth}{!}{%
\renewcommand{\arraystretch}{1.2}
\begin{tabular}{llllll}
\toprule
   Method & Epochs & Artificial data& Clean & AutoAttack     &  EoT AutoAttack  \\
\midrule
WRN-28-4 &400 & 1 M & \textbf{87.44}\%  & 58.28\%    &   -     \\
\emph{+ Prepro. 10/50} &  & &86.32\%        & \textbf{67.24}\%     &62.36\%   \\  
\midrule
WRN-28-10 &400 & 1 M  &\textbf{88.96}\%  & 61.60\%    &   -    \\
\emph{+ Prepro. 10/50} &  & &87.70\%        &68.56\%     &63.68\%   \\  
\emph{+ Prepro. 20/80} &  & &88.24\%        &\textbf{69.60}\%     &\textbf{67.2}\%   \\
\midrule
WRN-28-10 &2400 & 20 M  &90.12\%  & 64.40\%    &   -    \\  
\emph{+  Prepro. 20/80}&  & &\textbf{90.52}\%        &\textbf{73.08}\%     & 70.9\%   \\
\midrule
\midrule
SotA model \cite{bartoldson_adversarial_2024} &10000 &500 M &\textbf{93.68}\%  & 73.71\%    & -       \\
\midrule
WRN-82-12 \cite{bartoldson_adversarial_2024}  &3000 &150 M &93.04\%  & 71.41\%    & -       \\
\midrule
\emph{+ Prepro. 20/100}  &3000 &50 M &90.12\%  & \textbf{74.32}\%    & 73.00\%       \\
\bottomrule
\end{tabular}
}
\end{table*}
\subsection{Integration with state-of-the-art models}
\label{sec:results-SOTA}

In this second section, we evaluate how well our preprocessor integrates with SotA WRN models and compare their performance against top-ranked models on RobustBench \cite{croce2020robustbench}.
Specifically, we train WRN models of various sizes using adversarial training, both with and without our preprocessor, and evaluate their robustness using AutoAttack \cite{croce2020robustbench} as well as various other attacks. Note that given the computational requirements of such test, we use only the configuration noise then filter.
\begin{figure*}[ht!]
    \centering
    \includegraphics[width=0.9\linewidth]{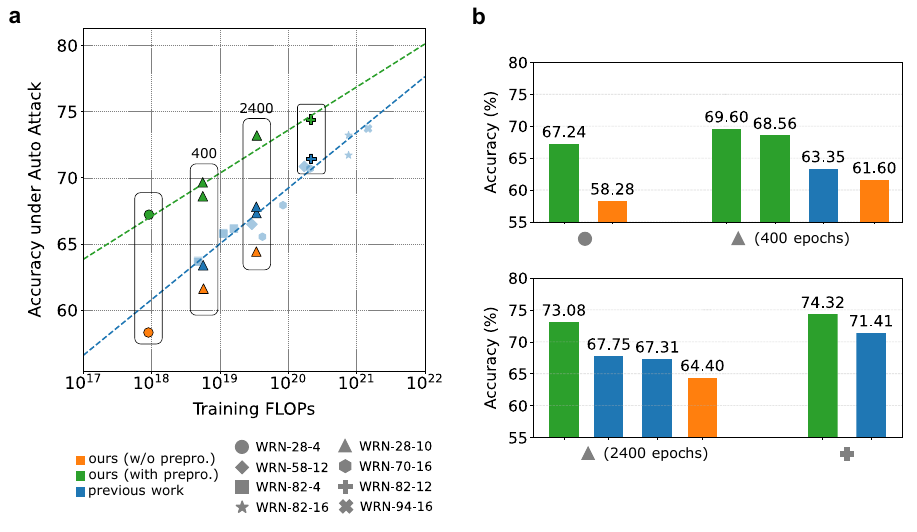}
    \caption{
    \textbf{a}) 
    Test accuracy (in \%) under AutoAttack in function of the number of training FLOPs (proportional to the number of training epochs, see Appendix \ref{sec:experimental-details}).
    Our WRN models trained with and without the preprocessor are shown in orange and green, respectively (see legend).
    Results from prior work \cite{wang_better_2023, cui2024decoupledkullbackleiblerdivergenceloss, bartoldson_adversarial_2024} are shown in blue.
    The results of the linear fit for models with our preprocessor are slope $a = 3.25 \pm 0.47$, intercept $b = 8.53 \pm 5.05$, and for prior work, slope $a = 4.21 \pm 0.23$, intercept $b = -14.98 \pm 4.56$ (see Appendix \ref{sec:experimental-details}).
    \textbf{b})
    Test accuracy of our WRN-28-4, WRN-28-10, and WRN-82-12 models, along with the corresponding models from \cite{wang_better_2023, cui2024decoupledkullbackleiblerdivergenceloss}, as highlighted by boxes in panel \textbf{a}.
    Models are color-coded according to the legend in panel \textbf{a}.
    }
    \label{SotA_result:f1}
\end{figure*}
\begin{figure*}[ht!]
    \centering
    \includegraphics[width=0.9\linewidth]{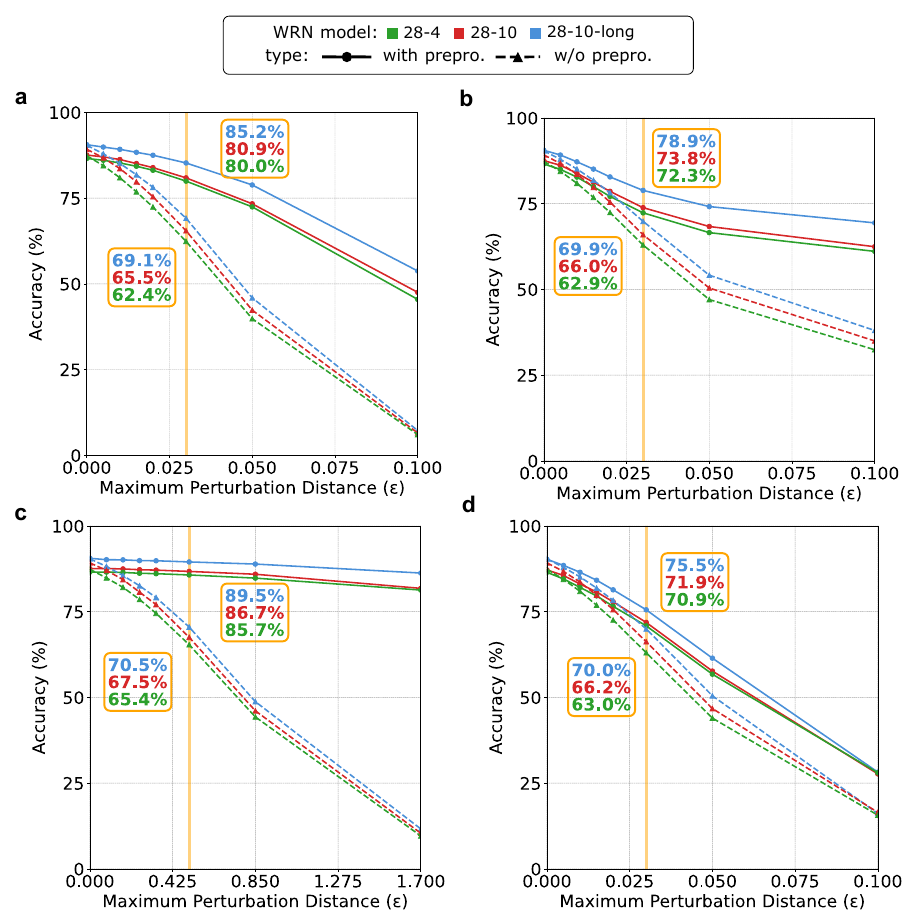}
    \caption{
    \textbf{a})
    Test accuracy (in \%) under the APGD-inf attack in function of the Maximum Perturbation Distance ($\epsilon$).
    Our WRN models without and with the preprocessor are plotted as solid and dashed lines, respectively (colors as indicated in the legend). 
    The \emph{WRN-28-10} model was trained for 400 epochs, while \emph{WRN-28-10-long} was trained for 2400 epochs.
    The orange vertical line marks the standard evaluation perturbation distances \cite{croce2020robustbench}, and we annotate each model’s accuracy at those points in the adjacent boxes.
    Details of the adversarial attacks used are provided in Appendix \ref{sec:experimental-details}.
    \textbf{b})
    Same as in \textbf{a} for the EoTPGD attack.
    \textbf{c})
    Same as in \textbf{a} for the APGD-L2 attack.
    \textbf{d})
    Same as in \textbf{a} for the TABPDA attack.
    }
    \label{SotA_result:f3}
\end{figure*}
\textbf{Integrating the preprocessor leads to improved adversarial robustness}.
In all our experiments, WRN models equipped with the preprocessor consistently achieve 6–9\% higher robust accuracy against standard AutoAttack, with only a minimal drop in clean accuracy (see Table~\ref{table:auto-attack}).
Against the computationally more expensive EoT variant of AutoAttack, our method achieves a 2–6\% improvement in robust accuracy compared to models without the preprocessor, with minimal clean accuracy loss (<1\%), and it even increases clean accuracy for the longer trained WRN-28-10. 
Our method scales effectively with the number of training epochs: training the WRN-28-10 model for 2400 epochs yielded 73.08\% robust accuracy, approaching the top-ranked models on RobustBench \cite{croce2020robustbench}.
Remarkably, our largest model tested the WRN-82-12 achieves a 0.6\% gain over the previous state-of-the-art without postprocessing \cite{amini_meansparse_2024} on the standard AutoAttack benchmark, while employing a model that is roughly half the size, needs 6x less synthetic data, and completes training in about one-third the number of epochs.

\begin{figure*}[ht!]
    \centering
    \includegraphics[width=\linewidth]{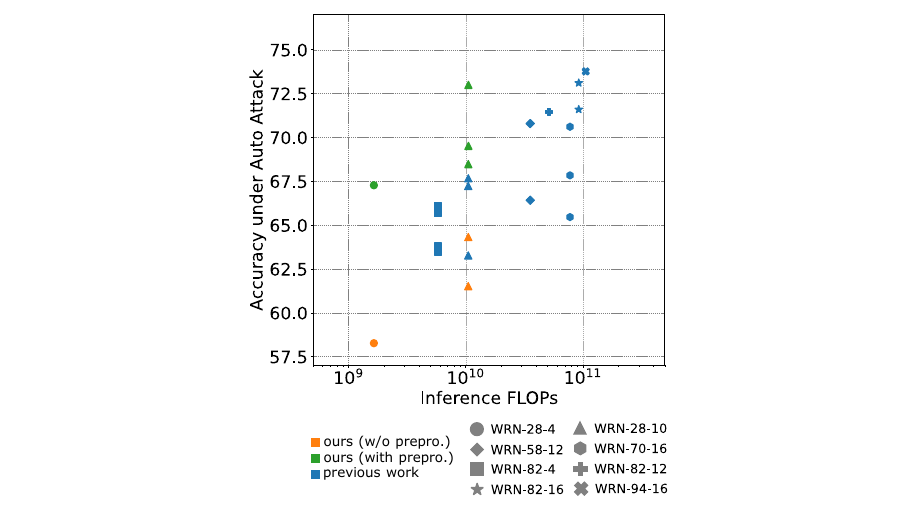}
    \caption{
    Test accuracy (in \%) under AutoAttack in function of the number of inference FLOPs, see Figure \ref{SotA_result:f1} in the main text.
    Our WRN models trained with and without the preprocessor are shown in orange and green, respectively (see legend).
    Results from prior work \cite{wang_better_2023, cui2024decoupledkullbackleiblerdivergenceloss, bartoldson_adversarial_2024} are shown in blue.
    }
    \label{suppfig:FLOPs-inference}
\end{figure*}

\textbf{The preprocessor scales efficiently with model size}.
Next, we analyze the relation between robust accuracy under AutoAttack  and the computational cost, measured in FLOPs, required for adversarial training.
Our comparison includes results from recent work \cite{bartoldson_adversarial_2024, wang_better_2023, cui2024decoupledkullbackleiblerdivergenceloss} listed in RobustBench \cite{croce2020robustbench}, including current and previous SotA methods.
When integrating our preprocessor into WRN models, we consistently observe higher robust accuracy compared to other models with similar training FLOPs, see Figure~\ref{SotA_result:f1}.
Specifically, our preprocessor improves the efficiency of robust accuracy relative to training cost: our models achieve the same target accuracy as competing methods while using only between 15\% and 50\% of their training FLOPs (see Appendix \ref{sec:experimental-details} for the mathematical derivation).
Additionally, our models achieve the highest accuracy relative to inference cost, matching the performance of models that require 6–9× more FLOPs at inference time (see Figure \ref{suppfig:FLOPs-inference}).

\textbf{The preprocessor is effective against a wide range of adversarial attacks}.
Finally, we evaluate the performance of our preprocessor against a wide range of adversarial attacks and perturbation strengths.
Specifically, we tested both the WRN-28-10 and WRN-28-4 models with the preprocessor against APGD-Linf \cite{croce_reliable_2020}, EoTPGD \cite{pmlr-v80-athalye18b}, and APGD-L2 \cite{croce_reliable_2020} attacks.
We also introduce a custom attack called TABPDA based on BPDA \cite{athalye_obfuscated_2018} which is tailored to our preprocessor.
TABPDA generates adversarial examples on images that have been processed only by the bilateral filter, omitting the Gaussian noise component (see Appendix~\ref{sec:experimental-details}).
As shown in Figure~\ref{SotA_result:f3}, our preprocessor improves robustness across all tested attacks, achieving at least a 5.5\% increase in robust accuracy at the most relevant perturbation level.
Additionaly, accuracy gains become more pronounced as the perturbation strength increases, particularly against L2 norm attacks, which were not used during training.

\section{Discussion}
In this work, we analyzed the effects of idealized de-noising filters and Gaussian noise on the geometry of adversarial attacks, revealing that both methods are effective at canceling different types of attacks. 
This leads us to hypothesize that a combination of both methods would cover a wide family of adversarial attacks.  
To test our theory by extending a standard CNN that uses a preprocessor combining bilateral filtering with Gaussian noise addition. As expected, the preprocessor yields supra-linear adversarial robustness compared to using either method on its own (see Section \ref{sec:results-adversarial-attacks}) with a minimal drop on clean accuracy.
Next, we test a combination of our preprocessor with SotA adversarially trained WRNs, benchmarking them using AutoAttack \cite{croce2020robustbench} as well as other adversarial attacks (see Section \ref{sec:results-SOTA}).
In all tested attacks, our preprocessor increases adversarial robustness compared to using adversarial training alone, even when we used a custom attack that targets our preprocessor explicitly. 
These gains are all the more relevant since our preprocessor induces minimal computational overhead, offering an efficient, yet powerful method to increase adversarial robustness.

\textbf{Concluding remarks.}
As our mathematical analysis of filters and noise revealed that they are effective against different underlying attack distributions, we can use our empirical observations to try to infer the underlying geometry of adversarial attacks. 
For example, the fact that bilateral filtering offers only modest gains in adversarial robustness would suggest that filamentary attacks are always present to some degree, and the supra-linear robustness given by the combined pre-processor indicates that some of those "filaments" can be rather thick.
However, those observations remain qualitative and untested, and more theoretical work should explore the shape of adversarial regions using tools from differential geometry.

Experimentally, we found a minor gap in the robust as well as the clean accuracies reported (blue) and reproduced (orange), even when using the exact same hyperparameters (see Figure \ref{SotA_result:f1}).
The cause of this gap is unknown to us, as the only difference is a slight combination of code from other papers \cite{bartoldson_adversarial_2024, wang_better_2023}, while the methods remain the same.
However, our results (green) still provide performance improvements beyond what has been reported (blue) for each model.
This could imply that using our preprocessor with an exact replica of the code from one of the other papers could further enhance resilience against AutoAttack.
Accordingly, the larger drop in clean accuracy we observe between our best model (WRN-82-12) and the WRN-82-12 from \cite{bartoldson_adversarial_2024} can be mostly attributed to slight training differences and the use of only a third of the artificial data.

Although our approach performs better for each WRN architecture and on Imagenet10, we observe diminishing returns: the improvement over the SotA WRN becomes smaller as model size and training resources increase (see Figure \ref{SotA_result:f1} and Appendix \ref{sec:experimental-details}) and for Imagenet10 only noise in the preprocessor is nearly as good as the combined preprocessor (see Table \ref{table:imagenet10}).
We consider two possible explanations for this effect.
First, humans and therefore artificial neural networks have an upper limit on classification accuracy under adversarial attacks around 90\% \cite{bartoldson_adversarial_2024}, suggesting a saturation effect for the WRNs.
Similarly for Imagenet10, if noise and bilateral filtering already work very well on their own, they may saturate the possible performance gain from their combination.
Second only for WRNs, large models with computationally expensive training may implicitly learn transformations similar to those applied by our preprocessor. 
In this case, our preprocessor would simply be a way to speed up training by applying these transformations explicitly from the start and thus accelerating convergence.
We further observed that integrating our preprocessor into SotA adversarially trained WRNs notably enhances robustness against adversarial attack types (\(L_2\)) and perturbation distances not seen during training (larger than \(8/255 \approx 0.03\)), see Figure \ref{SotA_result:f3}.
We believe that this gain may come from gradient masking \cite{athalye_obfuscated_2018} implicit in the bilateral filter, since our custom-made TABPDA (which bypasses gradient masking) does not show this notable gap at larger perturbation distances.
Yet, even under TABPDA our preprocessor still increases the robust accuracy, which suggests that a substantial portion of the accuracy gain arises from mechanisms not tied to obstructing adversarial example generation, but instead from genuinely increasing the intrinsic robustness of the overall model.

\textbf{Future work.}
Future work should explore other types of noise \cite{shi_defending_2022}, filtering techniques, and adversarial training methods, and their combinations, to find the most beneficial combination of the three.
A promising avenue to look for alternative filters and noise is to consider biologically inspired approaches.
In fact, the robustness of humans against adversarial attacks was an important component of our preprocessor design: bilateral filters are known to preserve shapes, an important human visual bias \cite{geirhos_imagenet-trained_2022}, and there is plenty of evidence of noise in early visual processing \cite{borghuis2009loss, freed2014synaptic}. 
Future work should consider further biological features and analyze their relevance to improve adversarial robustness.
Such work would not only offer benefits against adversarial attacks, but also help make CNNs closer to humans, an avenue of research that has been proven to be very fruitful in neuroscience \cite{richards2019deep,lindsay2021convolutional}
Further investigation into the differences in combinatorial gain across datasets could yield deeper insights into how filters, noise, and datasets interact in the context of adversarial robustness.
Furthermore, adapting adversarial training to include BPDA and EoT-based attacks could increase robust accuracy and training efficiency, since the PGD attack used in SotA adversarial training is hindered by gradient masking.
TABPDA may be a good candidate for this purpose as it combines both BPDA's and EoT's benefits, while it uses a similar amount of FLOPs as PGD. 
Moreover, new adversarial attacks that more effectively bypass our preprocessor could help in both gaining further insights into our preprocessor and in generating more robust models by integrating them into the adversarial training.
Such training attacks would have to be designed considering the randomness of the preprocessor without using EoT to stay efficient like TABPDA.
Finally, our work has focused on CNNs, but the theoretical insights underlying our approach are not specific to this architecture. 
This opens a promising direction for future research: applying and adapting our preprocessor to other models, such as vision transformers, to evaluate its effectiveness more broadly.

\\
\textbf{Code availability}: The code to reproduce the experimental results can be found at \url{https://github.com/Asinix13/simple-preprocessor-for-adversarial-robustnss-} 
{
    \small
    \bibliographystyle{ieeenat_fullname}
    \bibliography{main}
}

% WARNING: do not forget to delete the supplementary pages from your submission 
\clearpage
\setcounter{page}{1}
\maketitlesupplementary

\section{Imagenet10}
\label{sec:imagenet10}
We also did a short ablation study on imagenet10 \cite{liu2020imagenet10} to see if our preprocessor also works with other datasets see table \ref{table:imagenet10}.
First we clearly see that our preprocessor largely increases the accuracy for all adversarial attacks tested.
Further, bilateral filtering seems to be much more effective on Imagnet10 then on CIFAR-10.
However, we no longer see the supralinearity between the combination and its components.
The much stronger noise with a variance of 0.1 seems to be as effective as the combined preprocessor.
We also see that the bilateral filter seams to be much more effective on Imagenet10 then on CIFAR-10.
These to observations lead us to suspect that the missing supralinearity could come from saturation of robustness gain from the individual methods.
\begin{table*}[!ht]
\caption{
    Ablation study table, showing test accuracy (in \%) on the clean Imagenet10 dataset and under different adversarial attacks.
    We test standard CNN models \cite{tan_efficientnet_2020} trained without (first row) and with different defense methods.
    For \(L_{\infty}\) attacks, we set \(\epsilon = 0.015 \approx 4/255\), which is standard values in the literature \cite{das_keeping_2017,dziugaite_study_2016}.
    Hyperparameters for the adversarial attacks and the training are listed in the appendix Tables \ref{tabel:adv-hyperparams} and in appendix Section \ref{subsec:experimental-details-imagenet10}, respectively.
}

\label{table:imagenet10}
\vspace{11pt}
\centering
\resizebox{0.6\linewidth}{!}{%
\renewcommand{\arraystretch}{1.2}
{
\begin{tabular}{llllll}
\toprule\toprule
Method   & Clean & FGSM     &  \(L_{\infty}\) & EoT.  & TABPDA\\
% \bottomrule
%
% (result 1)
%
%
% & & & & & & &\\[1pt]
\bottomrule\bottomrule
Standard CNN \cite{tan_efficientnet_2020} &87.9\%   & 33.6\%      & 10.9\%     & 11.5\%         & 16.1\%             \\          
 & \(\pm\)0.9  & \(\pm\)3.1      & \(\pm\)1.7     & \(\pm\)1.6         & \(\pm\)1.6           \\
\midrule
+ Bil.  &86.7\%   &42.9\%      & 27.0\%     & 28.1\%         &30.7\%           \\
 & \(\pm\)0.5\  & \(\pm\)2.5      & \(\pm\)3.0     & \(\pm\)3.0       & \(\pm\)2.2            \\
+ Noise       &87.4\%   &45.3\%      &38.2\%     &36.1\%          &39.6\%         \\
& \(\pm\)1.0\  & \(\pm\)3.0      & \(\pm\)2.1     & \(\pm\)2.3     & \(\pm\)1.9            \\
+ Noise + Bil. &86.6\%   &46.0\%      &37.9\%     &37.1\%         &39.0\%           \\
& \(\pm\)0.6\  & \(\pm\)3.4      & \(\pm\)3.3     & \(\pm\)3.1     & \(\pm\)3.2            \\
+ Bil. + Noise &85.5\%   &44.8\%      &39.8\%     &39.1\%         &40.4\%           \\
& \(\pm\)0.9\  & \(\pm\)1.9      & \(\pm\)1.8     & \(\pm\)2.1     & \(\pm\)1.9            \\
\bottomrule
\end{tabular}
}}

\end{table*}

\section{Theory and derivations}
\label{sec:math}

In this section we present our theory suggesting that gaussian noise and filters have distinct mechanisms to enhance robustness against adversarial attacks, which result in them being useful against different types of attacks. We will first present our basic definitions and the conceptual framework that we will use, and then proceed to analyze the two approaches, and finally discuss some further arguments suggesting that both mechanisms should be combined. The theorems and arguments are illustrated in Fig.~\ref{fig:mathIllustration}

\subsection{Definitions and framework}

We consider images, where each one is a point $x$ in the space of possible images $\mathcal{X}$ with dimensions $D$ corresponding to the number of pixels. Within that space, there are manifolds of images corresponding to different image classes $C$. The various classes are disconnected, and there is a minimum distance $c$ between them. Furthermore, all unaltered images belong to one of those manifolds. Notice that this simplified definition precludes images that could be in two classes or inherently difficult to classify.

A machine learning model $f(x)$ is trained to take an image and identify which class does it belong to. The model generates a vector of dimension $C$, where each entry $f_c(x)$ is the probability of that image belonging to a specific class. We will assume that the model can classify the unaltered images $x$ with high certainty. This assumption implies that any ambiguity or mistakes in the classification comes from perturbing the image, not from lack of model capacity, training or inherent ambiguity.

Between any two classes there is a decision boundary $b(c_1,c_2)$ which determines the manifold on which the model changes which class has the highest probability.  This boundary is a $D-1$ connected manifold and is smooth with a finite sectional curvature \cite{fawzi2018empirical}. A perfect classifier would have  decision boundary is exactly between the two manifold classes $b^*(c_1,c_2)$. This perfect classifier is approximated by the model, but the approximation is not perfect, and there is an "adversarial volume" $V_a(x)$, which corresponds to the volume of between the real and the perfect classifier boundary, defined as
\begin{equation}
    V_a(x) = \int_{z\in \mathbf{B}(x,r)} \mathbf{1}\left[f(z)\neq h(z)=f(x)\right]dz.
\end{equation}

Now we proceed to define what we mean by adversarial attacks (AAs). We define them as an additive perturbation $a_x$ of a given image $x$ such that:
\begin{itemize}
    \item AAs have a bounded norm, meaning that $\|a_x\|\leq r$, and this norm is smaller than half the distance between classes $r< \delta/2$. Note that we did not specify which norm (as this depends on the AA considered).
    \item When added to the image, AAs cross the decision boundary and generate a misclassification.
    %\item We assume that if $x+a_x$ is misclassified, then $x+(1+\alpha)a_x$ is also misclassified for all $\alpha>0$. In other words, increasing the strength of an already successful adversarial attack does not make it disappear.
    \item The adversarial volume is very small. If adversarial attacks were frequent, it would be enough to add random noise to an image to find one (after a few samples). However, the generation of AA is based on the sophisticated machinery of deep learning optimizers. Thus, adversarial attacks must be a small subset of all possible image perturbations.
\end{itemize}
The relevant quantities for our analysis are explained graphically in Fig.~\ref{fig:mathIllustration}.

Finally, we note that adversarial attacks are intrinsically linked to the norm that is used to design them and to train against. For our derivations we used mostly concepts from Euclidean geometry, as they are the most well known and easy to understand, but other norms ($L_1$ or $L_\infty$) could yield similar results, albeit the bounds would be different because the notion of ball (sphere, in Euclidean spaces) is different.

\subsection{Comparing adversarial defenses}
The gist of our argument is that noise and filters are useful against different types of adversarial attacks. The key is then to explain what do we mean by different forms, and to come up with a way to make comparable attacks. This requires a new formalism that we introduce here.

Clearly, any defense mechanism depends on the type of attack that we are considering. To derive useful results that do not require explicit knowledge, we will aim to create upper bounds on the success rate of adversarial attacks under different defenses. However, such upper bounds require some constraint (otherwise we could simply consider the case where any attack would succeed). In this case, we will consider the constraint of a fixed $V_a(x)$.

Our approach would be to distribute the volume $V_a(x)$ in a way that minimizes the effectiveness of a specific defense mechanism. In doing so we will design the shape of the decision boundary that allows adversarial attacks, and in turn be able to compare how different attacks affect different defenses.

Before using this framework, we should note how it relates to adversarial training. In adversarial training, we identify one attack and use it to train the model, effectively pushing the decision boundary away from the image. This has the effect of reducing $V_a(x)$, although how is it reduced (meaning, which attacks are pushed back), heavily depends on the type of training used.

\subsection{Filters and filamentary attacks}

We begin by considering the use of filters as a defense mechanism. There is a wide variety of filters that have been proposed for a variety of functions. As some of those are not necessarily useful for image classification, we will restrict ourselves to de-noising filters such as JPEG, band-pass or bilateral filters. Even there, different filters can have different effects. To avoid these problems, we will consider an idealized denoising filter, formalized by a function $\varphi(x)$ and with the following properties:
\begin{itemize}
    \item The filters does not change the images in the class manifold. Implying that images that are not perturbed are not affected, meaning that
\begin{equation}
    \varphi(x) = x,\quad \forall x\in \mathcal{C}
\end{equation} 
This assumption is broken in practice by every filter, but it is a property that filter designers would try to achieve.
    \item The filter dampens down any deviation from the class manifold, meaning that 
\begin{equation}
\|\varphi(x+\epsilon) -\varphi(x)\|<\|\epsilon\| \ \forall x\in \mathcal{C} .
\end{equation}
In words, a filter dampens noise, where noise could be defined as any perturbation that is added to a "true" image (which we take as any point in the class manifold). It is worth noticing that filters rarely achieve zero noise (just improved signal-to-noise ratio). 
    \item The filter is a smooth function. While this is not strictly required for a filter (median filters for example, are not smooth), it is a valid statement for band pass and bilateral filters. It also will simplify our analysis significantly.
\end{itemize}

\subsection{Filters push adversarial attacks away from the image}

Consider now an adversarial attack on $f$ around an image that belongs to one of the class manifolds, and pass it through a filter. Since the norm of an adversarial attack is very small, we can take a Taylor approximation
\begin{equation}
    \varphi(x+ a_x) \approx \varphi(x) +J_{\varphi(x)} a_x = x +J_{\varphi(x)} a_x
\end{equation}
An attacker can take any attack on $f$ by applying the inverse function theorem on the Jacobian (if it exists), 
\begin{equation}\label{eq_filterInverse}
    \varphi(x+J_{\varphi(x)}^{-1} a_x) \approx x + a_x
\end{equation}
Since the filter dampens all deviations from the image, the eigenvalues of the Jacobian are semipositive, but smaller than one. By the inverse function theorem, the inverse of the Jacobian is expansive, meaning that
\begin{equation}
    \|J_{\varphi(x)}^{-1} a_x\| >    \| a_x\|.
\end{equation}
which implies that the filter forces any adversarial attack to have a bigger norm. More specifically, we can use the eigenvalues of the Jacobian $\lambda_{\varphi(x)}$ as bounds,
\begin{equation}\label{eq:filterBoundAA}
    \left(\lambda_{\varphi(x)}^{\min}\right)^{-1} \| a_x\|\geq \|J_{\varphi(x)}^{-1} a_x\| \geq \left( \lambda_{\varphi(x)}^{\max}\right)^{-1} \| a_x\|.
\end{equation}

This result could be generalized to non-smooth and non-invertible filters, but we opted for a simpler approach. This requires some justification:
\begin{itemize}
    \item First, if the Jacobian is not full rank, the inverse does not exist. If that is the case, the filter effectively eliminates one dimension. If the adversarial attack requires the use of such dimension, then the attack is automatically canceled. This agrees with the existing works showing that adversarial attacks are harder to find in lower dimensional spaces \citep{godfrey2023many}.
    \item Second, in principle we could make a similar argument without using the smoothness assumption, by simply considering the inverse of the filter and using the property that the distances are lowered by the filter. However, this would require us to track the range and domain of the filter and its inverse, which is a much more technical process than just using the Jacobian and the inverse function theorem.
\end{itemize} 

\subsection{Filamentary adversarial attacks against filters}\label{app:math_attackAgainstFilter}
Consider now the criteria for an adversarial attack that is able to surpass the filter. The main limitation of adversarial attacks is that they are bounded in norm, so we need to ensure that
\begin{equation}
    \|J_{\varphi(x)}^{-1} a_x\| <r
\end{equation}
Now, since adversarial attacks are simply points on the wrong side of the decision boundary, and this decision boundary is connected, we cannot simply put adversarial attacks close to $x$, but we need to link them to the decision boundary (because it is a connected surface). Furthermore, if the decision boundary has a finite curvature, we cannot create an infinitely-thin filament, but we must instead use a cylinder (a sphere in $D-1$ multiplied by a line). If we use the available volume in this manner,
\begin{equation}
    V_a(x) = S_{D-1}(c^{-1/2})l
\end{equation}
where $c$ is the maximum sectional curvature of the decision boundary (corresponding to the inverse of the square radius of curvature), and $S_{D-1}(z)$ is the volume of a sphere of radius $z$ and dimension $D-1$ given by
\begin{equation}
    S_{D-1}(c^{-1/2}) = \dfrac{\pi^{(D-1)/2}}{\Gamma(D/2+1/2)c^{D/2}}
\end{equation}
and which is multiplied by the cylinder length $l$.

We can thus isolate $l$ and derive the minimum norm of an adversarial attack as 
\begin{equation}
  \min\|a_x\| = r-  \alpha_{D-1}(x,c),
\end{equation}
where 
\begin{equation}
    \alpha_{D-1}(x,c) = \dfrac{V_a(x)}{S_{D-1}(c^{-1/2})}
\end{equation}
Combining this with the bound from Eq.~\ref{eq:filterBoundAA}, we conclude that there can be a successful attack if
\begin{equation}\label{eq:minDistFilter}
     \left(\lambda_{\varphi(x)}^{\min}\right)^{-1} \left(r-  \alpha_{D-1}(x,c)\right) \leq r,
\end{equation}
and if we phrase it in terms of the absolute norm, an adversarial attack would succeed if
\begin{equation}
    r \geq \dfrac{\alpha_{D-1}(x,c)}{(1-\lambda_{\varphi(x)}^{\min})}\end{equation}
which can be rephrased in terms of the adversarial volume by specifying that
\begin{equation}
    V_a(x) \geq (1-\lambda_{\varphi(x)}^{\min})rS_{D-1}(c^{-1/2}).
\end{equation}
Another point to highlight is that the maximum volume of adversarial attacks after the filter, is given by multiplying the adversarial by the eigenvalues of the Jacobian (which form the determinant), and scaling the cylinder to account for the length lost 
\begin{equation}
    V_{a}^{\varphi} \leq  V_a(x)|J_{\varphi}|\dfrac{r-\alpha_{D-1}(x,c)}{r}.
\end{equation}

\subsection{Noise probably avoids adversarial attacks}\label{app:noiseVsAA}

Now we consider the addition of noise in the input as a mechanism to defend against adversarial attacks. The gist of our argument is that adversarial attacks are rare on an neighborhood of the input, and therefore any unpredictable perturbation of the input of that would lead to an adversarial attack that is likely to fail.

We can add noise to the input $x+\varepsilon$, where $\sigma$ is a gaussian noise with a total variance of
\begin{equation}
    \sigma^2 = D \cdot\sigma_P^2,
\end{equation}
where $\sigma_P^2$ is the variance at each pixel.

Now consider an adversarial attack computed for the input $x$, $x+a_x$, on top of the noise we added before. The probability of an adversarial attack succeeding despite the noise then depends on the distribution of adversarial regions in the neighborhood. Formally,
\begin{equation}
    \text{Pr}\left[x+a_x+\varepsilon\in A \right] = \int_{\mathcal{X}}\text{Pr}\left[x+a_x+\varepsilon = z\right]\cdot \mathbf{1}\left[z\in A\right]dz
\end{equation}

\subsection{Spherical attacks against noise}\label{app:math_attackAgainstNoise}

We can take an upper bound on the previous probability by maximizing the probability of the attacked image and noise falling within the region of adversarial attacks
\begin{align}
    \text{Pr}\left[x+a_x+\varepsilon\in A \right] \leq \max_{A}\int_{\mathcal{X}}&\text{Pr}\left[x+a_x+\varepsilon = z\right] \cdot \mathbf{1}\left[z\in A\right]dz \\
    &s.t.\ \int_\mathcal{X} \mathbf{1}\left[z\in A\right] dz = V_a(x)(x)
\end{align}
since the noise is gaussian,
\begin{equation}
\text{Pr}\left[x+a_x+\varepsilon = z\right]= e^{\frac{-\|x+a_x-z\|^2}{2\sigma^2}}
\end{equation}
and thus the probability decreases monotonically with the distance from $x+a_x$. The highest probability is then when the attacks are in a sphere centered around $x+a_x$. From the volume we can infer the radius of this adversarial sphere,
\begin{equation}
    \rho(V_a(x))=  \sqrt[D]{V_a(x)\dfrac{\Gamma(D/2)}{\pi^{D/2}}} 
\end{equation}
which corresponds to the norm of $\varepsilon$ above which the noise will avoid the adversarial attack. We can then use this to estimate the probability of an adversarial attack succeeding,
\begin{align}
    \text{Pr}&\left[x+a_x+\varepsilon\notin A \right] \leq 
    \text{Pr}\left[\|\varepsilon\| >\rho(V_a(x))\right]\\
    &=  2\text{erf}\left(-\dfrac{\rho(V_a(x))}{\sigma}\right),
\end{align}
where erf is the cumulative gaussian function, and the $D$-th root term is the radius of a sphere of volume $V_a(x)$. Inverting using the complementary case,
\begin{equation}
    \text{Pr}\left[x+a_x+\varepsilon\in A \right] \leq  1-  2\text{erf}\left(-\dfrac{\rho(V_a(x))}{\sigma}\right).
\end{equation}

Notice that we have not mentioned where the sphere should be placed. Since there is a cost to moving the sphere close to $x$ (as the sphere needs to be connected to the optimal decision boundary), the natural choice would be to place the sphere at the boundary of the region where the attack is valid. This however, would imply that $x+a_x+\varepsilon$ can fall outside of the adversarial sphere. To make our bounds as tight as possible, we have implicitly assumed that the decision boundary is minimal outside of the adversarial sphere (in other words, the sphere of adversarial attacks is as isolated as possible).

\subsection{Attacks against noise and filters}\label{app:math_combination}

Taking the two optimal adversarial distributions against filters and noise, we notice that they are opposites: the best attack against a filter consists of a long and thin filament that approaches the unperturbed image as much as possible, while the best adversarial distribution against noise is a thick sphere placed at the boundary of the attack radius. 

This shows that the two defense mechanisms work against different families of attacks, and thus combining them would prevent different attacks. This sets the main theoretical motivation for our empirical approach. 

We can also go one step further and derive a bound for a combined defense, following the principles from earlier. The logic is simple: to overcome both defenses, we need to put the adversarial volume into a sphere (to overcome the noise) that needs to be close to the unperturbed image (to overcome the filter). 

The first step is to decide how much volume to allocate to moving the sphere close to the image. Our requirement is to be within the radius that the filter would not be able to push back, which we can obtain from Eq.~\ref{eq:minDistFilter} as
\begin{equation}
    c_a = r(1-\lambda_{\varphi(x)}^{\min})
\end{equation}
which multiplied by the section of the cylinder yields
\begin{equation}
    V_a^{\varphi}(x) = r(1-\lambda_{\varphi(x)}^{\min})S_{D-1}(c^{-1/2}).
\end{equation}
Notice that if $V^{\varphi}_a(x)\geq V_a(x)$, there is not enough volume, and the best option is the filamentary attack. If that is not the case, the leftover volume can be used to make a sphere. 

However, here we must consider two possibilities: either the noise is injected before the filter or afterwards. 
If the noise is injected before the filter, we can take the noise as is and consider a sphere with volume
\begin{equation}
    V_a^{e}(x)= V_a(x) - V_a^{\varphi}(x),
\end{equation}
yielding a bound similar to what we obtained in Eq.~\ref{eq:boundOnNOise},
\begin{equation}
     \text{Pr}\left[\varphi(x+a_x+\varepsilon)\in A \right] \leq  1-  2\text{erf}\left(-\frac{\rho(V_a^{e}(x))}{\sigma}\right).
\end{equation}
However, if the noise is injected after the filter, the sphere will be scaled by the inverse of the Jacobian (because the effective noise is on image space), yielding
\begin{equation}
    V^{\varepsilon}_a(x) = (V_a(x) - V_a^{\varphi}(x))|J_{\varphi(x)}|^{-1}
\end{equation}
which in terms similar to Eq.~\ref{eq:boundOnNOise} gives
\begin{equation}
     \text{Pr}\left[\varphi(x+a_x+\varepsilon)\in A \right] \leq  1-  2\text{erf}\left(-\frac{\rho(V_a^{\varepsilon}(x))}{\sigma}\right).
\end{equation}
Notice that the effect of the Jacobian scaling relies heavily on the nature of the filter and the attack. In general $|J_{\varphi(x)}|^{-1}<1$ due to the contractive nature of the filter, thus applying the filter before the noise is better for adversarial attacks. However, it is also worth considering that applying a filter after the noise has the added benefit of reducing the noise, and therefore increasing the clean accuracy (or conversely, allowing for a higher noise). 

This is leads to a subtle effect: applying the filter then adding noise might be better to defend against adversarial attacks, but makes the system more vulnerable to the preprocessor noise. In turn, this decrease in clean accuracy might permeate to adversarial conditions, lowering the accuracy of the system under attacks (but due to the vulnerability to noise, not to lower adversarial resilience). 

Thus, our theory cannot decide whether the filter should be applied before or after the noise in terms of adversarial resilience, but clean accuracy would benefit from the former order. 

\section{Detailed description of the preprocessor components}
\label{sec:filtering-and-noise-methods}
\label{appendix:a}
In this section, we provide a detailed description of the image processing methods used in our preprocessor.
Mostly we first add zero-mean Gaussian noise independently to each color channel of every pixel, and then apply multiple bilateral filters to the resulting image.
We also test changing the order of noise and filtering, which we specifically indicate.
In our implementation, the preprocessor is integrated as an additional layer that can used be during both training and inference.

\textbf{Gaussian noise.}
When Gaussian noise is added to an image, each pixel is perturbed by a random value drawn from a zero-mean Gaussian distribution:
\[
I_{noisy}(p) = I(p) + \mathcal{N}(0, \sigma^2) \,,
\]
where \(\mathcal{N}(0, \sigma^2)\) denotes a random value from a Gaussian distribution with mean 0 and variance \(\sigma^2\).  
Increasing the variance \(\sigma^2\) results in stronger noise.

\textbf{Bilateral filter.}
The bilateral filter \cite{tomasi_bilateral_1998} is a non-linear filter that extends Gaussian filtering by incorporating both spatial and intensity (or color) information.  
Unlike the Gaussian filter, which considers only the spatial distance between pixels, the bilateral filter also accounts for the difference in intensity or color, making it edge-preserving.
For a given pixel \( p \), the filtered output is computed as:
\[
I_{\text{filtered}}(p) = \frac{1}{W_p} \sum_{q \in \Omega} I(q) \cdot G_{\sigma_s}(||p - q||) \cdot G_{\sigma_r}(|I(p) - I(q)|) \,,
\]
where \( \Omega \) is the set of neighboring pixels,
\( G_{\sigma_s}(||p - q||) \) is the spatial Gaussian kernel based on the Euclidean distance between pixels \( p \) and \( q \),
\( G_{\sigma_r}(|I(p) - I(q)|) \) is the range kernel (a Gaussian function of the intensity or color difference), and \( W_p \) is a normalization factor ensuring the weights sum to one.
The spatial kernel \( G_{\sigma_s} \) ensures that nearby pixels have more influence, while the range kernel \( G_{\sigma_r} \) reduces the influence of pixels with different intensities or colors, preserving edges.
This filtering is both filter-specific and pixel-specific, as the kernel adapts dynamically based on the local image structure.
The bilateral filter smooths images while preserving edges, reducing texture without blurring important boundaries.  
Increasing either \( \sigma_s \) (spatial standard deviation) or \( \sigma_r \) (range standard deviation) increases the overall smoothing effect.  
However, increasing \( \sigma_r \) specifically reduces sensitivity to intensity or color differences, allowing the filter to smooth across stronger edges.

\textbf{Other filters.}
We also evaluated other types of filters, which we list below for completeness.
\begin{itemize}
    \item 
    A Gaussian filter is a linear filtering technique used to smooth images.  
    It works by convolving the image with a Gaussian kernel, which weights neighboring pixels based on their Euclidean distance from the center pixel,
    \[
    I_{\text{filtered}}(p) = \frac{1}{W_p} \sum_{q \in \Omega} I(q) \cdot G_{\sigma_s} (||p - q||) \,.
    \]
    Here, \( p \) is the coordinate of the current pixel, \( q \) denotes a pixel in the neighborhood \(\Omega\),  \( W_p \) is a normalization factor ensuring the weights sum to one, \( I(x) \) is the intensity value at position \( x \), and 
    \( G_{\sigma_s} (||p - q||) \) is the spatial Gaussian kernel,
    \[
    G_{\sigma_s} (||p - q||) = \frac{1}{2\pi\sigma_s^2} e^{-\frac{||p - q||^2}{2\sigma_s^2}} \,,
    \]
    where \( \sigma_s \) is the spatial standard deviation, where higher values of \( \sigma_s \) result in stronger smoothing.
    Because Gaussian blur treats all pixels equally, regardless of their intensity values, it tends to remove fine textures and smooth out edges, leading to a loss of shape information. 
    Nonetheless, it has been shown to increase the shape bias in CNNs \cite{hermann_origins_2020}.
    \item 
    Median filters are non-linear filters that replace each pixel’s value with the median of the intensity values in its local neighborhood.  
    The operation is defined as:
    \[
    I_{\text{filtered}}(p) = \text{median}(\{I(q) \mid q \in \Omega\}) \,.
    \]
    This results in a filter that is both pixel-specific and data-dependent, where only one position in the kernel, that is the median, is effectively used.
    This makes the filter highly adaptive to local image content.
\end{itemize}

\clearpage

\section{Experimental details}
\label{sec:experimental-details}
In our implementation, the preprocessor is integrated as an additional layer that can used during both training and inference.
We use adapted functions from the Kornia library \cite{riba_kornia_2019}, a computer vision toolkit built on PyTorch, to implement the preprocessing steps. 
These include the optional addition of pixel-wise Gaussian noise and the application of bilateral filters to the input data.
\subsection{Experimental setup and training for the CIFAR-10 experiments}
The hyperparameters used for the different preprocessors we tested are listed in Table~\ref{tabel:preproc-hyperparams}.
All experiments were conducted using the CIFAR-10 dataset \cite{cifar_10}, which consists of 10 classes, each containing 6,000 images, for a total of 50,000 training images and 10,000 test images. The images are color images with a resolution of 32x32, stored as Python floats in the range [0, 1]. 
For Fast, we standardize the data, which moves it outside the original input range. We then perform a hyperparameter search to find the optimal settings for both training and preprocessing parameters.
A full implementation of the preprocessing module and training script is available in our supplementary code repository \url{https://github.com/Asinix13/simple-preprocessor-for-adversarial-robustnss-}.

\textbf{Experimental details for the standard CNNs}.
The standard CNN used is EfficientNet-B0 \cite{tan_efficientnet_2020}.
All models, except for the adversarially trained "Fast is Better Than Free" (Fast) \cite{wong_fast_2020}, were trained for 80 epochs on fully shuffled, unperturbed training data.
Fast, was trained and tested on standardized data before any perturbations were applied the rest was done on non standardized data.
To train Fast, we adopted both the methodology and hyperparameters from Wong et al.'s \cite{wong_fast_2020} GitHub repository.
The hyperparameters for standard training, as well as for Fast, are presented in Table \ref{tabel:training-hyperparams}.
The hyperparameters for the preprocessor can be found in Table \ref{tabel:preproc-hyperparams}.
We conduct these experiments using NVIDIA GeForce RTX 3090 GPUs. 
Runs are performed on a single GPU and take up to 4 hours, depending on the batch size and the other tasks running on the same GPU.
The adversarial attacks test runs, excluding C\&W, take around 5 hours per run. For the C\&W attacks, we use 2 GPUs, and each run takes approximately 4 hours.
The exact timing depends on the additional workload on the GPUs and on the batch sizes.
Additionally, roughly two weeks of using 4-8 GPUs were dedicated to hyperparameter tuning and testing the implementations.
CIFAR-10 requires approximately 177 MB of storage, the standard models take up around 20 MB.

\begin{table*}
\caption{
Training hyper-parameters for the standard CNN experiments.
}
\label{tabel:training-hyperparams} 
\vspace{11pt}
\centering
\begin{tabular}{ll}
\toprule
 & Hyperparameters\\
\midrule
     Standard training  &opt. = Adam;  lr. = 1e-3;  epochs = 80;  batch size = 64\\
\midrule
     Fast training  &opt. = Sgd;  lr. min. = 0;  lr. max = 0,05;  delta init. = random  \\ 
      &momentum = 0,99;  weight decay =  5e-4 \\
      &lr. schedul = cyclic; early stopping = true; loss scale = 1.0   \\
      &epochs = 80;  batch size = 64; opt. level = O2\\
      &master weights = true; gamma = 0,1 \\
      &epsilon = 8/255;  alpha = 10/255; PGD alpha = 2/255;\\
      &PGD attack iter. = 5;  PGD restarts = 1\\
      
\bottomrule
\end{tabular}

\end{table*}

\begin{table*}[ht!]
    \caption{
    Preprocessing hyper-parameters for the ablation study. Bilateral filter amount training is only used during the vanilla CNN training. For all preprocessors, the following conditions are consistent: Gaussian noise always has a mean of zero; the sigma space for Bilateral filters is fixed at 10; the sigma range for the initial Bilateral filter is 0.1, and the filter size is 5. For subsequent Bilateral filters, the filter size is reduced to 3.
    }
    \vspace{11pt}
\label{tabel:preproc-hyperparams} 
\centering
\begin{tabular}{lllll}
\toprule
Preprocessor & Gauss var. & Bil. fil. amount train. & Bil. fil. amount inf. & Sigma range\\
\midrule
     Bil. fil. (b)  &- & 10 & 10 & 0,05\\
     Gauss noi. (g) &0.032 &- &- &- \\ 
     (g) and (b) &0.032 & 10 & 10 & 0,05\\
     Best (g) and (b) &0.032 & 10 & 50 & 0,05\\
\midrule
\midrule
JPEG & qualitiy = 25 & & & \\
\bottomrule

\end{tabular}

\end{table*}

\textbf{Experimental details for the WRN models.}
We use WRNs \cite{zagoruyko_wide_2017} using Swish activation function \cite{hendrycks_gaussian_2023}.
For the SotA adversarial training we use the TRADES method \cite{zhang_theoretically_2019} implemented by Wang et al. \cite{wang_better_2023}, which was then slightly adapted by Bartoldson et al. \cite{bartoldson_adversarial_2024} and we then slightly adapted further for our own purpose.
Our code is therefore based on both GitHub repositories; Wang: \url{https://github.com/wzekai99/DM-Improves-AT} and Bartoldson \url{https://github.com/bbartoldson/Adversarial-Robustness-Limits}.
We further use Wang et al.'s through elucidating diffusion model \cite{Karras2022edm} generated artificial data, which can be downloaded from their repository.
The exact hyperparameters for every model and the adversarial training we used , are presented in Table \ref{table:WRN-train}. 
The hyperparameters for the preprocessor can be found in Table \ref{table:WRN-preproc}.
We conduct all experiments using NVIDIA GeForce RTX 3090 or NVIDIA Quadro RTX 6000 GPUs.
For all these experiments, expect for those using the WRN-82-12, we use 4 RTX 3090 together.
For all experiments using the WRN-82-12 we use 8 RTX 6000 together.
The WRN-28-4 take around 7 hours to train, the WRN-28-10 around 7 days and the WRN-82-12 around 20 days all on the respective hardware we used.
\begin{table*}
\caption{
Adversarial training hyper-parameters for the WRN experiments.
}
\label{table:WRN-train} 
\vspace{11pt}
\centering
    \begin{tabular}{ll|ll}
        \toprule
        \textbf{Hyperparameter} & \textbf{Value} & \textbf{Hyperparameter} & \textbf{Value} \\
        \midrule
        normalize & false & adv eval freq & 50 \\
        beta & 5 & lr & 0.2 \\
        weight decay & 0.0005 & scheduler & cosinew \\
        nesterov & true & clip grad & null \\
        attack & linf-pgd & attack eps & 8/255 \\
        attack step & 0.00784 & attack iter & 10 \\
        keep clean & false & mart & false \\
        unsup fraction & 0.7 & seed & 1 \\
        consistency & false & cons lambda & 1.0 \\
        cons tem & 0.5 & LSE & false \\
        ls & 0.1 & clip value & 0 \\
        CutMix & false & tau & 0.995 \\
        unfix N batches per epoch & false & better sampler & false \\
        EDM 50 amount & null & pct start & 0.025 \\
        config & null & robustblock & false \\
        one epoch & false & batch sizes validation 128 & \\
        \bottomrule
    \end{tabular}

\end{table*}

\begin{table}[ht!]
    \caption{
    Preprocessing hyper-parameters for the WRNs experiments.}
    \vspace{11pt}
\label{table:WRN-preproc} 
\centering
\begin{tabular}{ll}
\toprule
  & Hyperparameter\\
\midrule
     Gauss mean &0 5\\
     Gauss var.& 0.032 \\ 
\midrule
\midrule
    First bil. filter & \\
\midrule
     Sigma space &10\\
     Sigma range &0.1\\
     Size &5x5\\
\midrule
\midrule
    Subsequent bil. filter & \\
\midrule
    Sigma space &10\\
     Sigma range &0.05\\
     Size &3x3\\
\bottomrule

\end{tabular}

\end{table}

\subsection{Experimental setup and training for the Imagenet10 experiments}
\label{subsec:experimental-details-imagenet10}
The Imagenet10 dataset \cite{liu2020imagenet10}  is a subset of the normal Imagenet dataset \cite{deng2009imagenet} containing 10 classes with 1300 images each.
We use the same 80/20 train/test data split for each experiment.
For the preprocessors components in our Imagnet10 experiments, we use a variance of 0.1 for the Gaussian noise, 10x bilateral filters, a sigma range of 10 for the first filter and 0.5 for the following filters, and a sigma space of 50 for all filters.
The kernel size for the first filter is 5 and for the following filters it is 3.
For training the EfficientNet-B0 \cite{tan_efficientnet_2020} we use in this experiments, we use the Adam optimizer with a learning rate of 1e-3 and a weight decay of 1e-6, a batch size of 128 and train over 60 epochs.
The training as well as the testing runs are all done on a single NVIDIA GeForce RTX 3090.
\subsection{Adversarial attacks and noise robustness}
In this section, we describe the types of adversarial attacks used, our implementation of a true averaged BPDA attack, and the experiments conducted to evaluate noise robustness.

\textbf{Description of the adversarial attacks}.
We employed seven different adversarial attacks to evaluate the robustness of our models: FGSM \cite{goodfellow_explaining_2015}, APGD with two different perturbation bounds \cite{croce_reliable_2020}, EoTPGD (EoT) \cite{pmlr-v80-athalye18b}, TABPDA our own adapted BPDA \cite{athalye_obfuscated_2018}, Auto Attack \cite{croce_reliable_2020} and the C\&W attack \cite{carlini_towards_2017}.
To save space, the two APGD attacks will also be referred to as \(L_{2}\) and \(L_{\infty}\) based on their respective norms in Table \ref{table:full-tabel}.
FGSM, APGD, EOTPGD, and BPDA are gradient-based attacks, bounded by a maximum perturbation distance \(\epsilon\). 
The C\&W attack, in contrast, minimizes the \(L_{2}\) distance of the perturbation while optimizing for an adversarial example.
FGSM conducts a single gradient-based step, while the other gradient-based attacks perform iterative steps following the PGD method.
APGD automatically searches for optimal attack parameters, while EOTPGD accounts for model randomness by averaging over multiple perturbation samples to mitigate inherent Gaussian noise. 
These methods can also be grouped by the distance metric used to measure perturbation. The C\&W attack and one version of APGD utilize the \(L_{2}\) distance, whereas the other attacks rely on the \(L_{\infty}\) distance. 
By testing various norms and attack strategies, we obtain a more comprehensive understanding of the model's robustness.
For \(L_{\infty}\) attacks, we tested with \(\epsilon\) values of 0, 0.005, 0.01, 0.015, 0.02, 0.03, 0.05, and 0.1. 
For \(L_{2}\) APGD attacks, we tested with \(\epsilon\) values of 0, 0.085, 0.17, 0.255, 0.34, 0.51, 0.85, and 1.7.
For the C\&W attack, we used \(c\) values of 1, 1.2, 1.4, 1.6, 1.8, and 2.
The other attack-specific hyperparameters are listed in Table \ref{tabel:adv-hyperparams}.
For all attacks except TABPDA, we used the Torchattacks library \cite{kim2020torchattacks}.
\begin{table}
    \caption{
    Adversarial attacks hyper-parameters.
    }
    \vspace{11pt}
\label{tabel:adv-hyperparams} 
    \centering
    
    \begin{tabular}{ll}
    \toprule
    
        Adv. attack & Hyper-parameters \\
    \midrule
    
         FGSM  & -  \\
    \midrule
         APGD \(L_{\infty}\)  & steps = 10;  restarts = 1; loss = ce\\
          & seed = 0; EoT itr. = 1; rho = 0.75 \\

    \midrule
         APGD \(L_{2}\)  & steps = 10;  restarts = 1; loss = ce\\
          & seed = 0; EoT itr. = 1; rho = 0.75 \\
    \midrule 
         EoTPGD & alpha = 2/255 ; steps = 50 ; EoT itr.  = 2\\
         & random start = true\\
    \midrule
         TABPDA & learning rate = 0.5 ; max itr. =  10\\
    \midrule
        C\&W & kappa = 0;  steps = 50; lr. = 0.01\\
    \bottomrule
    \end{tabular}

\end{table}
In Table \ref{table:full-tabel}, we set \(\epsilon = 0.03 \approx 8/255\) for \(L_{\infty}\) attacks and \(\epsilon = 0.51\) for \(L_{2}\) attacks \cite{das_keeping_2017,dziugaite_study_2016}. 
Furthermore, for the C\&W attack, we set \(c = 1.8\) and \(c = 2\) as the strongest attacks for the standard CNN and the SotA network, respectively.

\textbf{Our custom True Average Backward Pass Differentiable Approximation (TABPDA)}.
TABPDA (True Average BPDA) is a custom attack designed to bypass gradient masking caused by stochastic preprocessors. It generates adversarial examples by ignoring the stochastic component and instead using the true average output of the preprocessor, ensuring uninterrupted gradient flow.
We adapt BPDA by replacing the full preprocessed image (which includes noise) with a deterministic version that is filtered only by the bilateral filters from the original preprocessor. Since the added noise has zero mean, this filtered image represents the true expected output of the full preprocessor, making it a valid approximation for gradient-based attacks.
In this way, TABPDA can be seen as a hybrid of BPDA and EoT: rather than averaging gradients over multiple stochastic samples as in EoT, it uses the known true average directly. This makes TABPDA significantly more efficient, but it requires knowledge of the preprocessor's exact average behavior.

\textbf{Robustness against Gaussian noise}.
To evaluate the robustness of the model against Gaussian noise, we apply zero-mean noise with variances of 0, 0.001, 0.002, 0.004, 0.008, 0.012, 0.016, 0.02, and 0.03 to all three color channels of each pixel. For the standard CNN (without the Fast variant) we calculate the robust accuracy over the last 10 training steps. For all other experiments, we test at the final training step.
The Gaussian noise, labeled "Gauss" in Table \ref{table:full-tabel}, corresponds to a variance of 0.03.

\subsection{Calculation of the preprocessor FLOPs}
Here, we estimate the floating point operations (FLOPs) of our preprocessor.
For a single application of bilateral filtering on an RGB image, the computational complexity is given by:
\begin{equation}
    O(\alpha*K^2*H*W)
\end{equation}
where \( K \) is the kernel size, \( H \) and \( W \) are the image height and width, and \( \alpha \) is a constant that depends on the choice of color distance norm:  
\(\alpha = 32\) for the \( \ell_2 \) norm and \(\alpha = 33\) for the \( \ell_1 \) norm.
In our implementation, adding Gaussian noise requires approximately 50 to 100 times fewer FLOPs than applying bilateral filtering 10 times, and at least \(\sim\)25,000 times fewer FLOPs than the smallest tested model (see Table \ref{table:flop}).  
For this reason, we omit the FLOPs contribution from Gaussian noise in our total FLOPs calculations.
To compute the total FLOPs used for adversarial training, we follow the procedure described in \cite{bartoldson_adversarial_2024}:  
we multiply the inference FLOPs per image by 27 (the cost of one adversarial training step), and then multiply that by the total number of training images.
\begin{table*}[!ht]
\caption{Floating point operation for one forward pass of one a CIFAR-10 image (32x32x3) through a full WRN or through parts of our preprocessor.
}
\label{table:flop}
\vspace{11pt}
\centering
\renewcommand{\arraystretch}{1.2}
\begin{tabular}{llllll}
\toprule
   & WRN-24-4 & WRN-24-10 & WRN-82-12 & 10x Bil. Filter  &  Gauss. Noise  \\
\midrule
Forward FLOPs &1.69 G   & 10.49 G & 51.81 G & 3.58 M    & 67.68 K       \\

\bottomrule
\end{tabular}
\vspace{10pt}
\end{table*}

\subsection{Comparison of FLOPs efficiency via linear fit analysis}
\begin{figure*}[!ht]
\centering
\includegraphics[width=\linewidth]{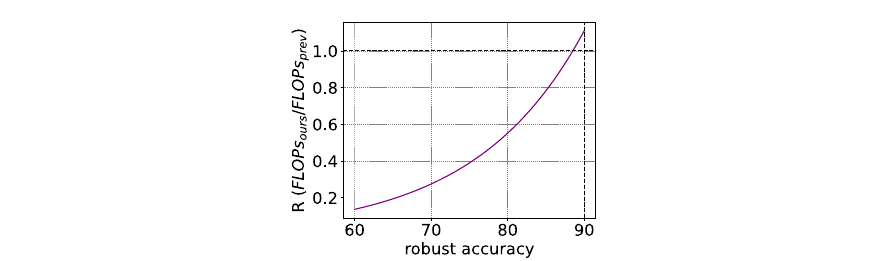}
\caption{
The ratio between the FLOPs required by prior work and by our method as a function of a target robust accuracy. 
We set 90\% accuracy as the human-level robustness threshold as suggested in \cite{bartoldson_adversarial_2024})
}
\label{suppfig:ratio}
\end{figure*}

In Figure \ref{SotA_result:f1}, we present a linear fit describing the relationship between test accuracy under Auto Attack and the total number of FLOPs during training, considering both our WRN models and prior works.
The linear fit follows the form
$A = a\log(\text{FLOPs}) + b$, where $A$ denotes the test accuracy. 
In particular, we obtain the following linear fits:
\begin{itemize}
    \item $a_{prev} = 4.21 \pm 0.23$,  $b_{prev} = -14.98 \pm 4.56$
    \item $a_{ours} = 3.25 \pm 0.47$,  $b_{ours} = 8.53 \pm 5.05$
\end{itemize}
where the subscripts 
$prev$ and $ours$ refer to the linear fits corresponding to previous work and our models, respectively.
Using these two linear models, we compare the FLOPs required by each method to achieve a given test accuracy $A_*$, calculated as
\begin{equation}
    \text{FLOPs}_* = \exp\left(\frac{A_*  - b}{a}\right) \,.
\end{equation}
This leads to the following ratio between the FLOPs required by prior work and by our method at the same accuracy level,
\begin{equation}
\resizebox{0.85\columnwidth}{!}{$
    R(A_*) = \exp\!\left[{A_*}\!\left(\frac{1}{a_{\text{ours}}} - \frac{1}{a_{\text{prev}}}\right)
    + \left(\frac{b_{\text{prev}}}{a_{\text{prev}}} - \frac{b_{\text{ours}}}{a_{\text{ours}}}\right)\right]
$}
\end{equation}
where $R(A_*) < 1$ means that our methods requires less training FLOPs than the previous models to achieve the given test accuracy $A_*$, see Figure \ref{suppfig:ratio}.

\clearpage
\section{Supplementary figures}
\label{sec:sup-mat}
\begin{table*}
\caption{
    Test accuracy (in \%) on the clean CIFAR10 dataset and under different adversarial attacks.
    We test standard CNN models \cite{tan_efficientnet_2020} trained without (first row) and with different defense methods.
    The best results are highlighted in bold.
    For \(L_{\infty}\) attacks, we set \(\epsilon = 0.03 \approx 8/255\), and for \(L_{2}\) attacks, we set \(\epsilon = 0.51\), which are standard values in the literature \cite{das_keeping_2017,dziugaite_study_2016}.
    For the C\&W attack, we set \(c = 1.8\)  for the standard CNN as the strongest tested attacks.
    The variance of the Gaussian noise perturbation is set to the highest tested value, 0.03.
    Hyperparameters for all methods and the adversarial attacks are listed in the appendix Tables \ref{tabel:preproc-hyperparams} and \ref{tabel:adv-hyperparams}, respectively.
}

\label{table:full-tabel}
\vspace{11pt}
\centering
\renewcommand{\arraystretch}{1.2}
{
\begin{tabular}{llllllll}
\toprule\toprule
Method   & Clean & FGSM     &  \(L_{\infty}\) & EoT.  & \(L_{2}\)& C\&W & Gauss\\
% \bottomrule
%
% (result 1)
%
%
% & & & & & & &\\[1pt]
\bottomrule\bottomrule
Standard CNN \cite{tan_efficientnet_2020}           &\textbf{74.5}\%   & 3.5\%      & 0.2\%     & 0.2\%         & 1.3\%     & 0.6\%  & 31.9\%      \\          
& \(\pm\)2.4  & \(\pm\)2.1      & \(\pm\)0.4     & \(\pm\)0.5         & \(\pm\)0.6     & \(\pm\)0.6  & \(\pm\)3.2      \\
\hline
+ Bil.                 &69.0\%   &10.0\%      & 1.0\%     & 1.2\%         &11.9\%     & 0.5\%   &60.3\%   \\
 & \(\pm\)2.2\  & \(\pm\)1.2      & \(\pm\)0.4     & \(\pm\)0.5        & \(\pm\)1.4     & \(\pm\)0.3  & \(\pm\)1.8\      \\
 
+ Noise       &68.5\%   &22.8\%      &25.5\%     &12.0\%          &49.6\%     &43.0\%  & \textbf{61.3}\%  \\
& \(\pm\)1.7\  & \(\pm\)1.3      & \(\pm\)1.2     & \(\pm\)0.9     & \(\pm\)1.4    & \(\pm\)1.0  & \(\pm\)1.2      \\

+ Noise + Bil. &67.9\%   &33.9\%      &36.5\%     &18.9\%         &58.5\%     &47.2\%   &59.6\%   \\
& \(\pm\)1.4\  & \(\pm\)1.7      & \(\pm\)1.5     & \(\pm\)1.4     & \(\pm\)1.7    & \(\pm\)1.0  & \(\pm\)1.8\      \\

+ JPEG  \cite{das_keeping_2017,dziugaite_study_2016, guo_countering_2017}           &70.0\%   &12.7\%      & 0.6\%     & 0.8\%     & 2.7\%     & 0.4\%    &46.9 \% \\
& \(\pm\)4.6\  & \(\pm\)2.4      & \(\pm\)0.4     & \(\pm\)0.4      & \(\pm\)0.7    & \(\pm\)0.4  & \(\pm\)2.8\      \\

+ Fast \cite{wong_fast_2020}   &71.0\%   &43.7\%      &\textbf{48.1}\%     &\textbf{40.3}\%     &58.6\%     &20.9\%  &57.5\%   \\
& \(\pm\)1.3\  & \(\pm\)1.4      & \(\pm\)1.5     & \(\pm\)1.3  & \(\pm\)1.6    & \(\pm\)3.8  & \(\pm\)2.5\      \\

+ Best Prepro.                 &67.4\%   &\textbf{43.9}\%      &47.3\%     &30.8\%       &\textbf{65.1}\%     &\textbf{56.6}\%   &60,5\%  \\
& \(\pm\)1.4\  & \(\pm\)1.8      & \(\pm\)1.4     & \(\pm\)1.5  & \(\pm\)1.5    & \(\pm\)1.3  & \(\pm\)1.4\      \\
\bottomrule

\end{tabular}
}
\end{table*}

% \subsection{Filter and noise comparison}
\begin{figure*}[ht!]
    \centering
    \begin{subfigure}[t]{0.7\linewidth}
        \includegraphics[width=\linewidth]{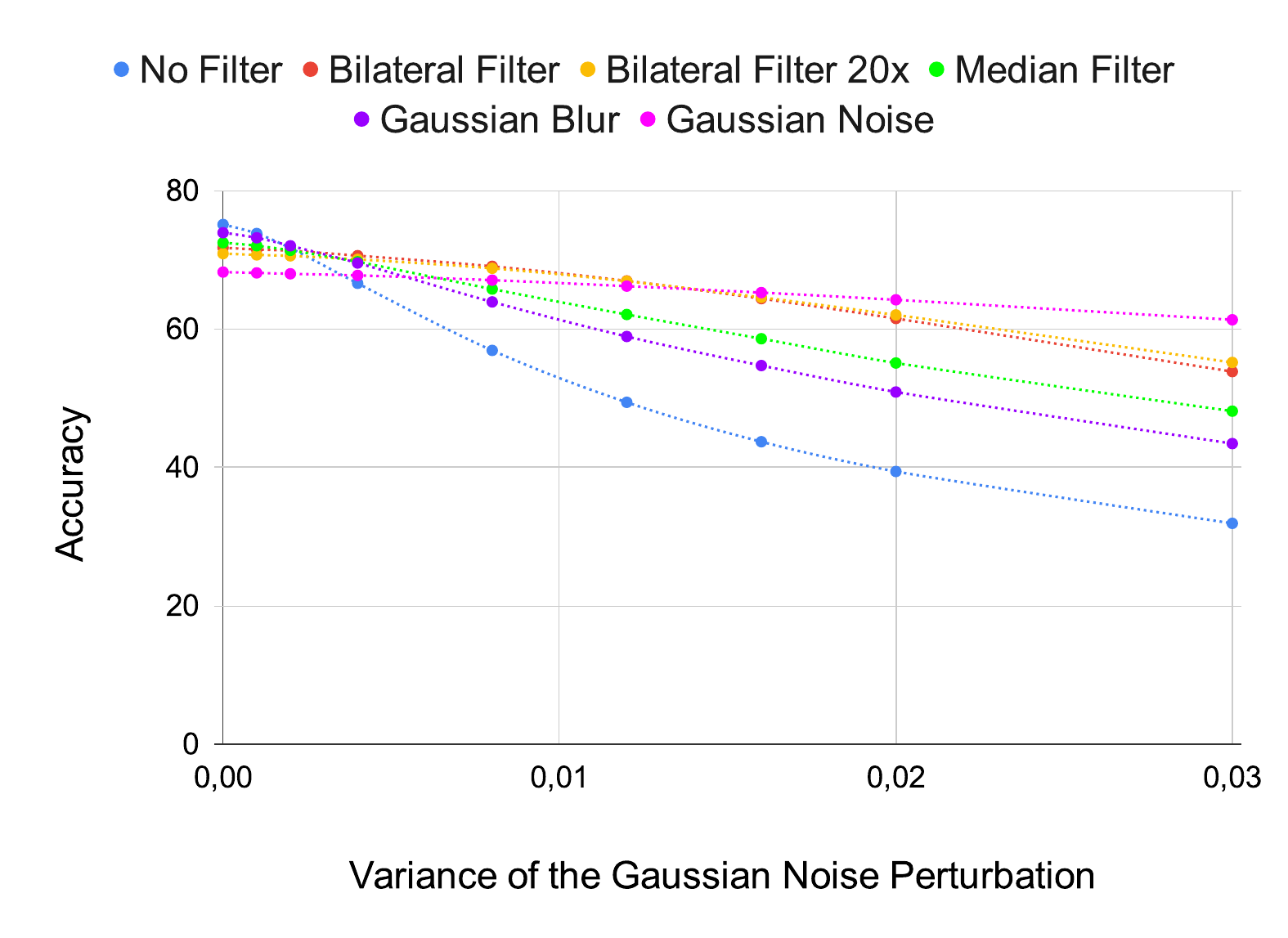}
        \caption{Different filters applied once per image.}
        \label{fig:noise_comp_a}
    \end{subfigure}
    \hfill
    \begin{subfigure}[t]{0.7\linewidth}
        \includegraphics[width=\linewidth]{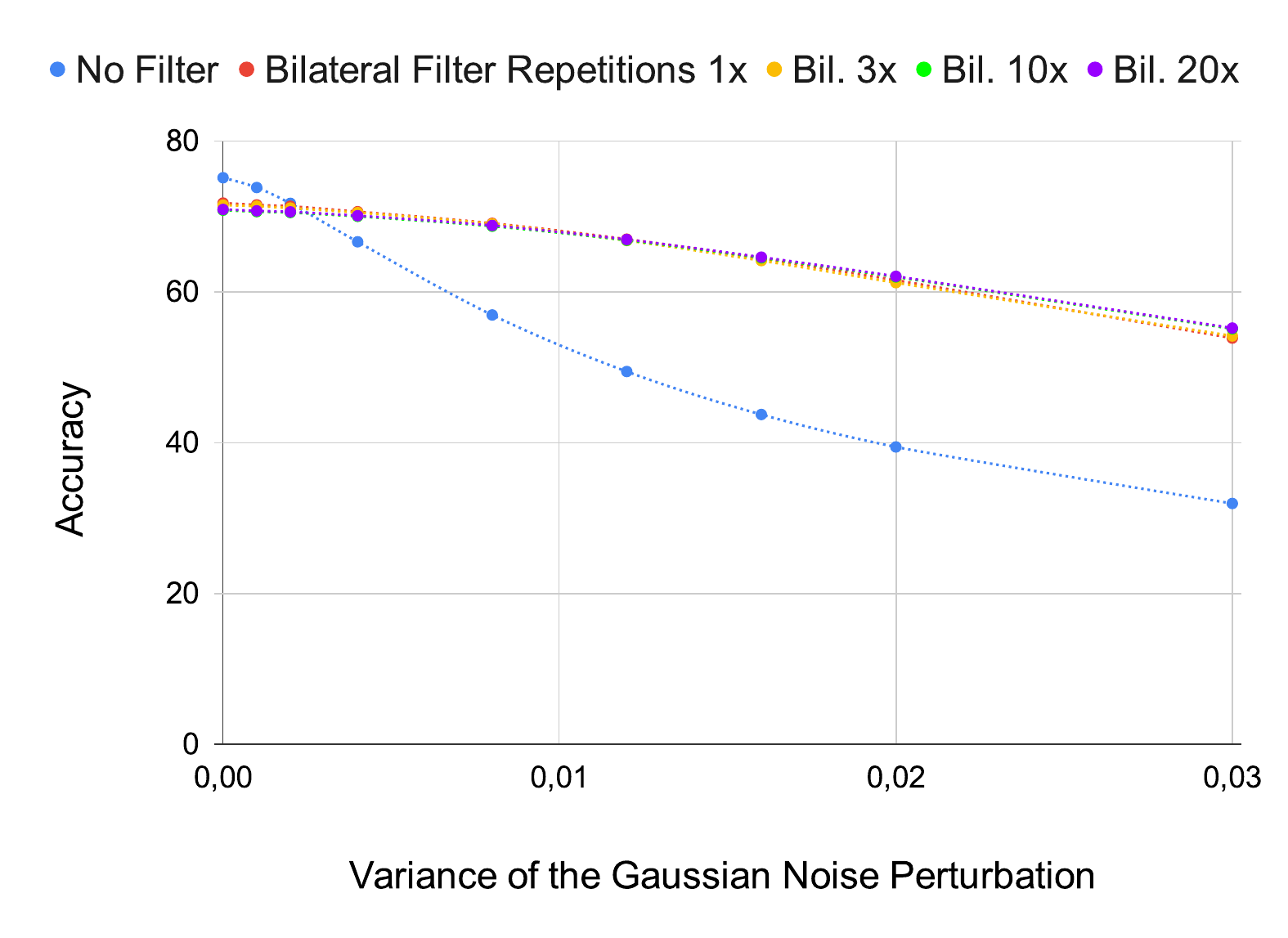}
        \caption{Iterative application of bilateral filter.}
        \label{fig:noise_comp_b}
    \end{subfigure}
    \caption{Comparison against Gaussian noise. Average accuracy for a range of Gaussian noise variances.
    The plot a compares different filters applied once per image, while plot b shows results of iterative bilateral filtering.
    Multiple iterations do not noticeably improve performance compared to single application.}
    \label{fig:noise_comp}
\end{figure*}

\begin{figure*}[ht!]
    \centering
    \begin{subfigure}[t]{0.7\textwidth}
        \includegraphics[width=\linewidth]{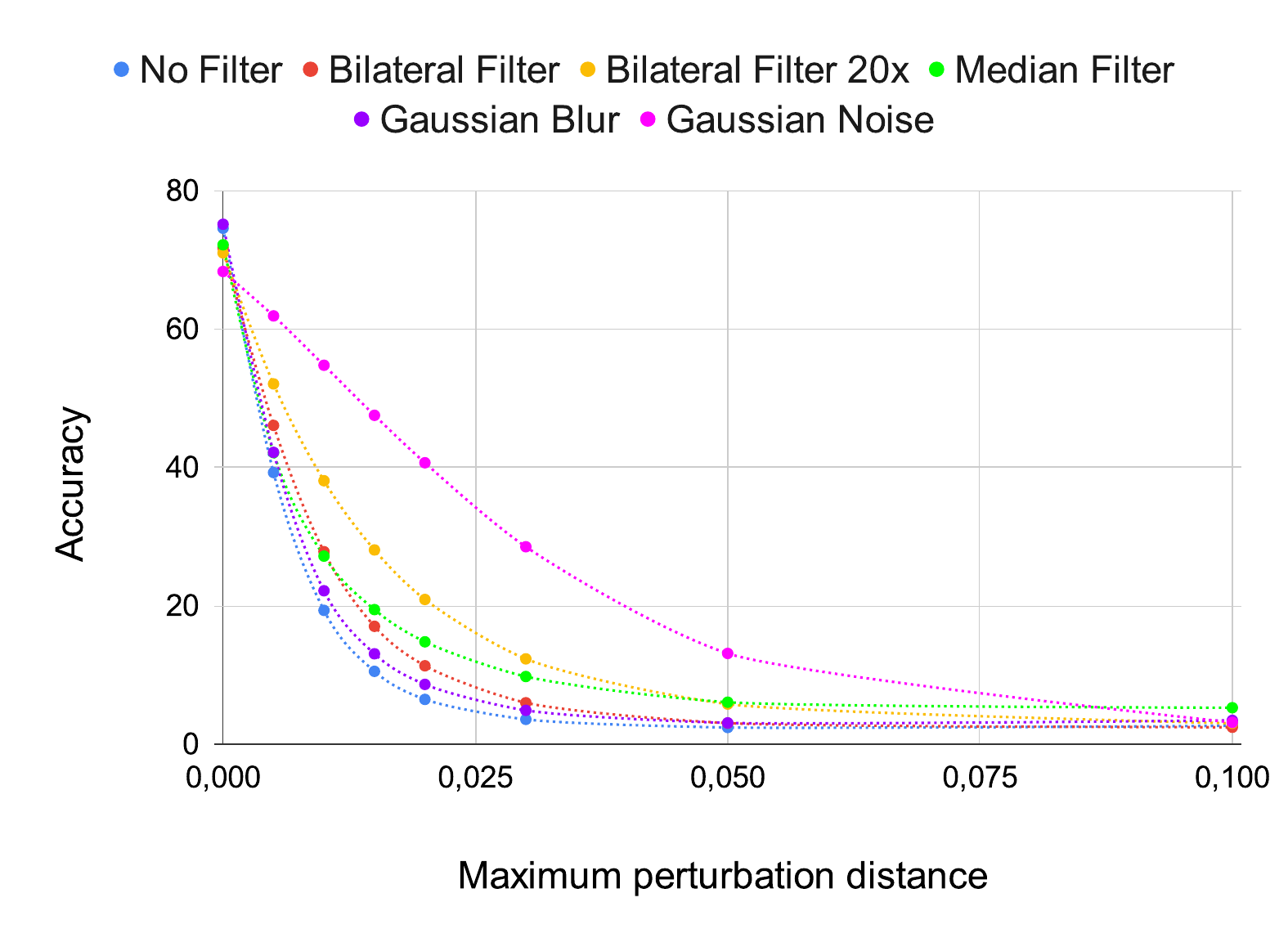}
        \caption{Different filters applied once per image. Gaussian noise achieves high accuracy under FGSM attacks. 
        For weaker attacks, the bilateral filter (20×) performs second best, while the median filter degrades more gracefully than the bilateral filter.}
        \label{fig:fgsm_comp_a}
    \end{subfigure}
    \hfill
    \begin{subfigure}[t]{0.7\textwidth}
        \includegraphics[width=\linewidth]{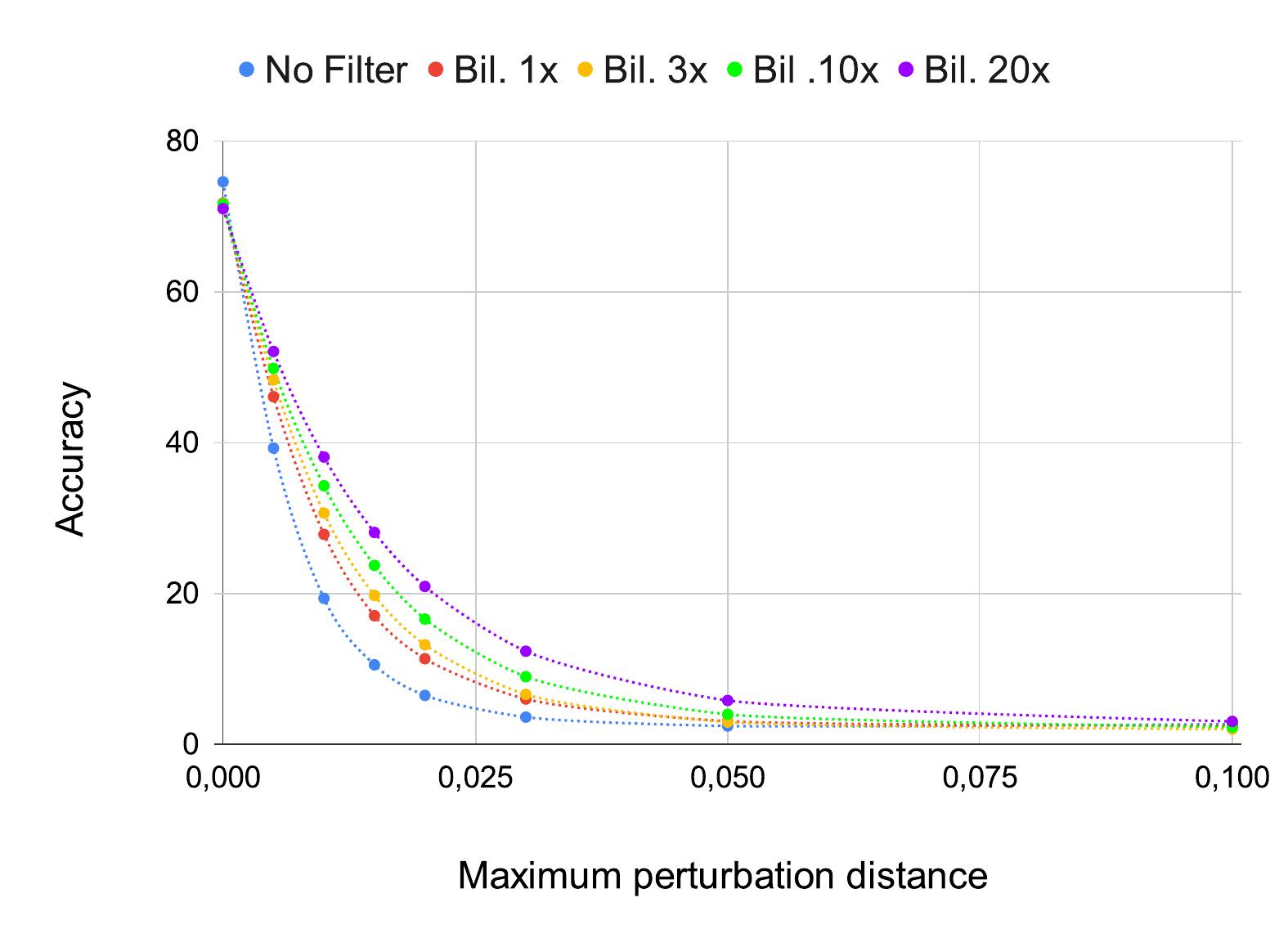}
        \caption{Iterative application of the bilateral filter. Repeated filtering increases robustness against FGSM attacks.}
        \label{fig:fgsm_comp_b}
    \end{subfigure}
    \caption{Comparison against FGSM adversarial attacks. Average classification accuracy across different noise and filtering strategies.}
    \label{fig:fgsm_comp}
\end{figure*}

% \subsection{The preprocessor scales with model size during inference}
%

% \subsection{Accuracy of different methods against adversarial attacks on Standard CNN}

%%%%%%%%%%%%%%%%%%%%%%%%%%%%%%%%%%%%%%%%%%%%%%%%%%%%%%%%%%%%%%%%%%%%%%%%%%%%%%%
% \subsection{Accuracy of different methods against Gaussian noise}
\label{sup-mat:e4}
%%%%%%%%%%%%%%%%%%%%%%%%%%%%%%%%%%%%%%%%%%%%%%%%%%%%%%%%%%%%%%%%%%%%%%%%%%%%

% --- First part of the figure ---
\begin{figure*}[ht!]
    \centering
    
    % Subfigures 1 and 2
    \begin{subfigure}[b]{0.8\textwidth}
        \centering
        \includegraphics[width=\linewidth]{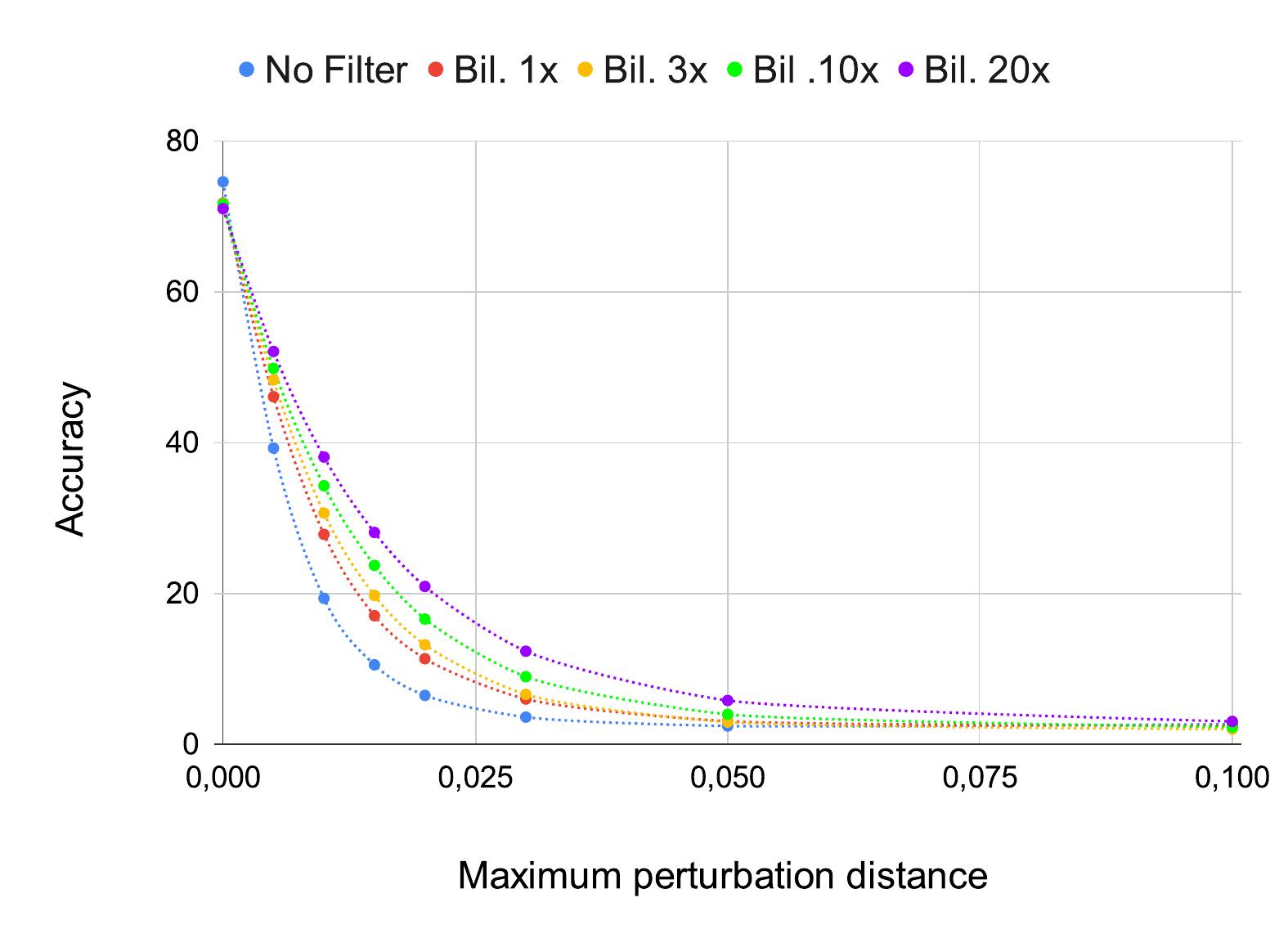}
        \caption{Accuracy under FGSM varying the bilateral filter repetitions without Gaussian noise addition. Sigma range is 0.05.}
        \label{fig:sens_1}
    \end{subfigure}

    \vspace{0.5cm}

    \begin{subfigure}[b]{0.8\textwidth}
        \centering
        \includegraphics[width=\linewidth]{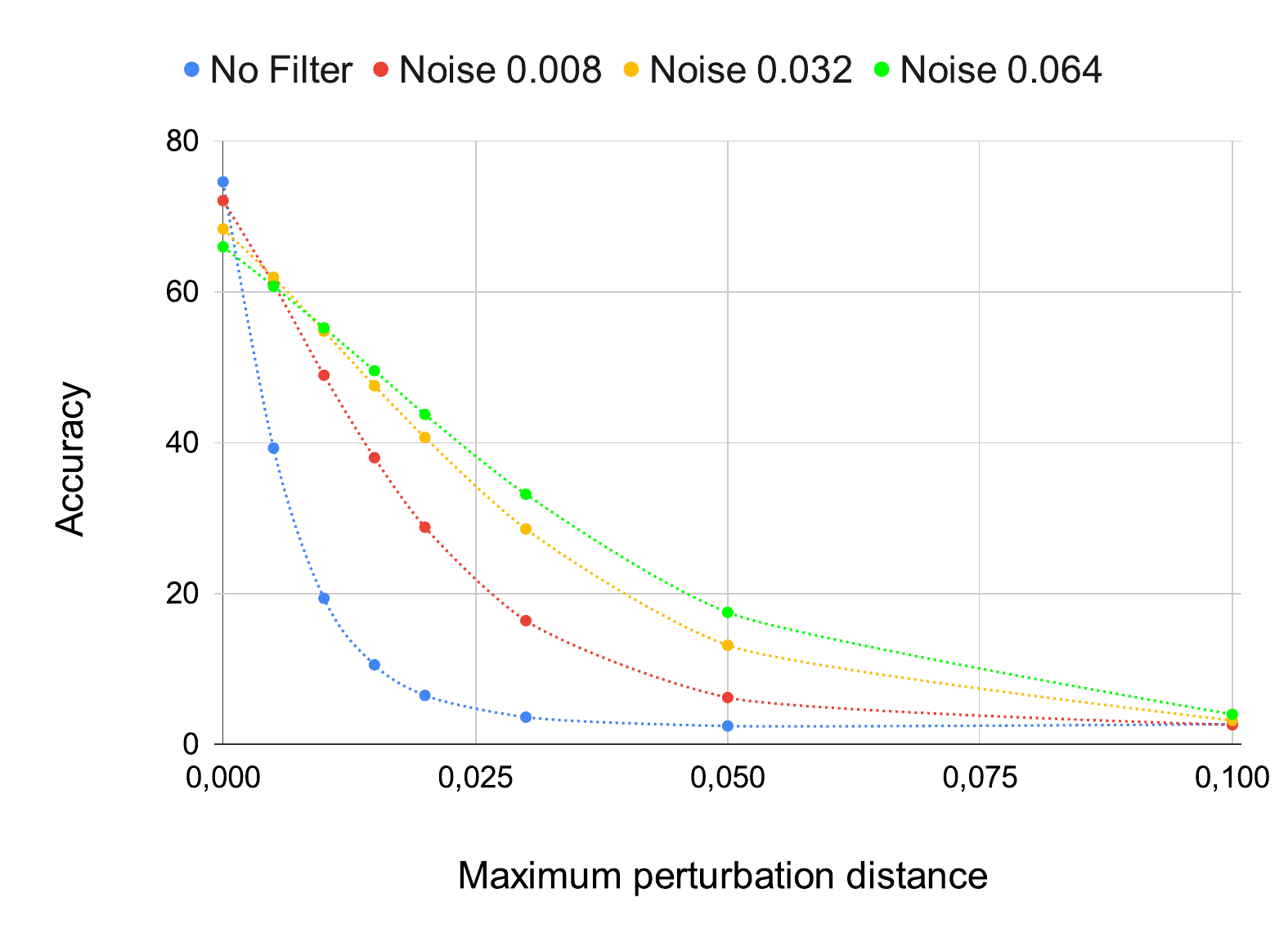}
        \caption{Accuracy under FGSM varying the Gaussian noise variance repetitions without bilateral filtering. The number next to Noise is the variance.}
        \label{fig:sens_2}
    \end{subfigure}

    \caption{Sensitivity study (Part 1). Continued on next page.}
    \label{fig:sensitivity}
\end{figure*}

% --- Force page break before continuing ---
\clearpage

% --- Continued on next page (subfigures 3–4) ---
\begin{figure*}[ht!]\ContinuedFloat
    \centering

    \begin{subfigure}[b]{0.8\textwidth}
        \centering
        \includegraphics[width=\linewidth]{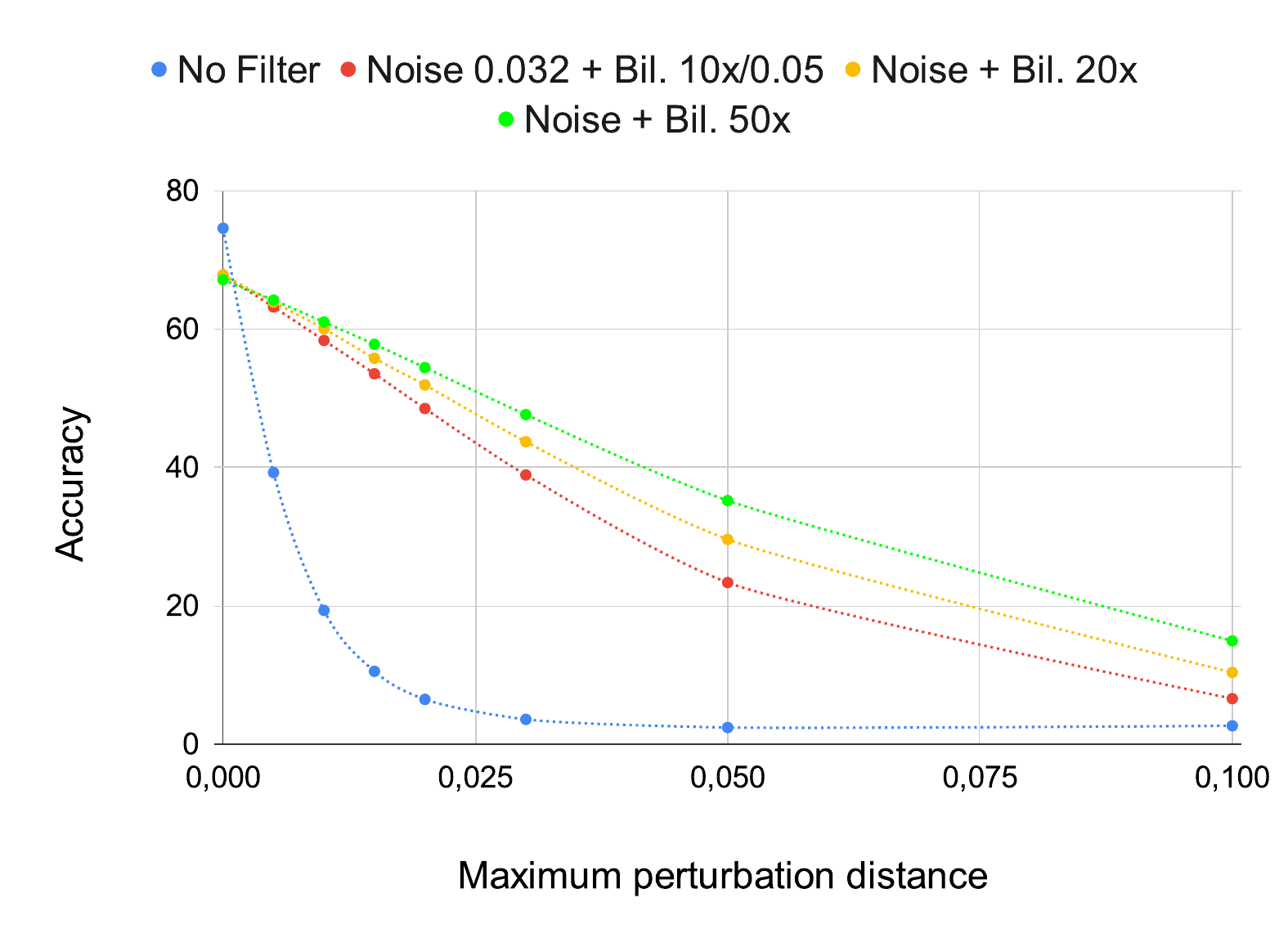}
        \caption{Accuracy under FGSM varying the bilateral filter repetitions with Gaussian noise addition. Sigma range is 0.05 and the Gaussian noise variance is 0.032.}
        \label{fig:sens_3}
    \end{subfigure}

    \vspace{0.5cm}

    \begin{subfigure}[b]{0.8\textwidth}
        \centering
        \includegraphics[width=\linewidth]{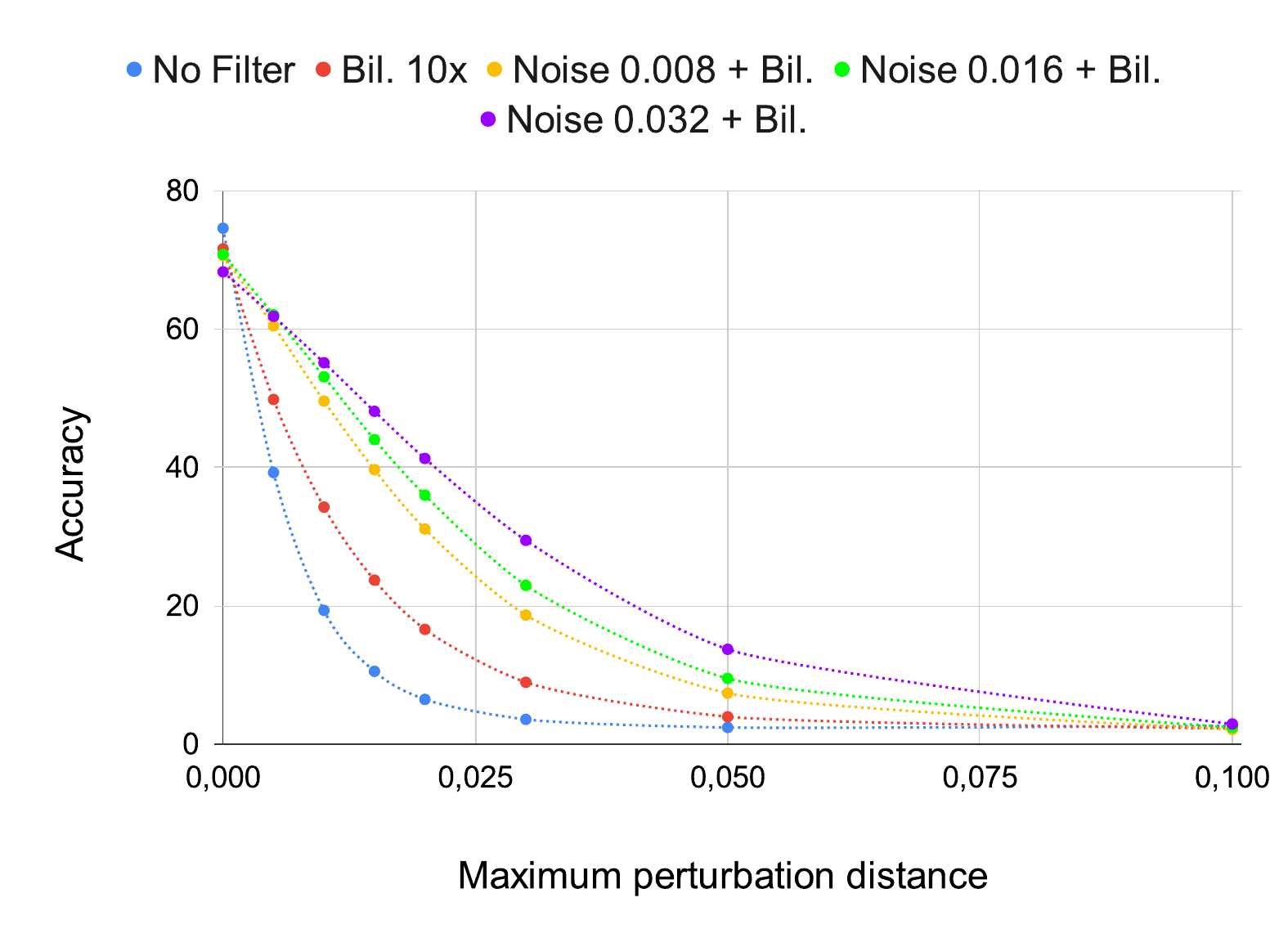}
        \caption{Accuracy under FGSM varying the Gaussian noise variance repetitions with bilateral filtering. The number next to Noise is the variance. Bilateral filter repetition amount is 10x and sigma range is 0.05.}
        \label{fig:example_4}
    \end{subfigure}

    \caption{Sensitivity study (Part 2). Continued on next page.}
\end{figure*}

% --- Force another page break for the last subfigure ---
\clearpage

% --- Final part (subfigure 5 only) ---
\begin{figure*}[ht!]\ContinuedFloat
    
    \centering

    \begin{subfigure}[b]{0.8\textwidth}
        \centering
        \includegraphics[width=\linewidth]{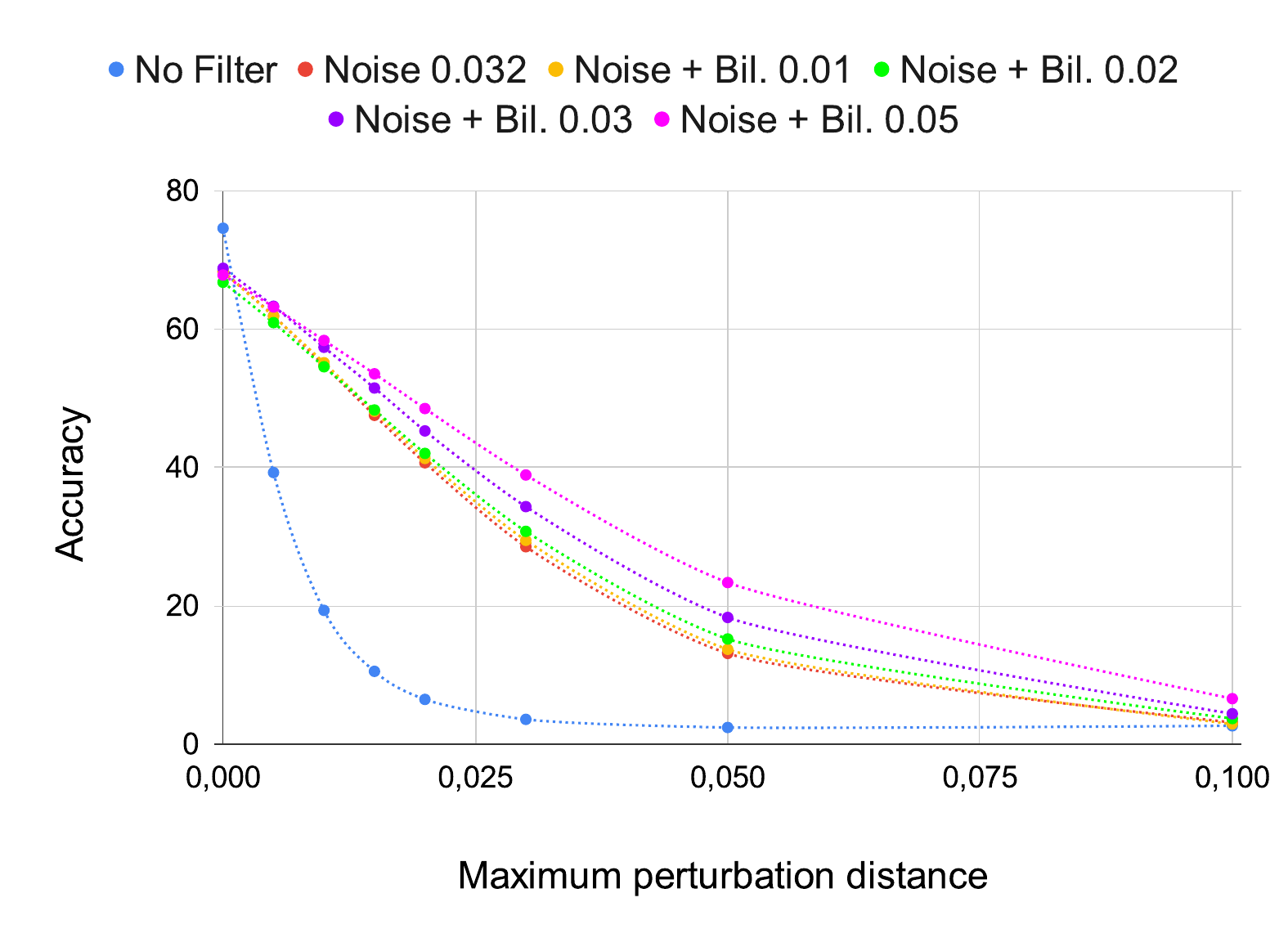}
        \caption{Accuracy under FGSM varying the bilateral filters sigma range with Gaussian noise addition. The number next to Bil. is the sigma range. Bilateral filter repetition amount is 10x and the Gaussian noise variance is 0.032.}
        \label{fig:example_5}
    \end{subfigure}
    
    \caption{Sensitivity study (final part). For each figure if there is a bilateral filter used, the sigma space is 10, the sigma range for the first bilateral filter is 0.1, the size of the first bilateral filter is 5x5 and the size for the following bilateral filters is 3x3.}
    
\end{figure*}

\end{document}